\documentclass[final]{IEEEtran}

\usepackage{scalerel}
\usepackage{tikz}
\usetikzlibrary{svg.path}
\definecolor{orcidlogocol}{HTML}{A6CE39}
\tikzset{
  orcidlogo/.pic={
    \fill[orcidlogocol] svg{M256,128c0,70.7-57.3,128-128,128C57.3,256,0,198.7,0,128C0,57.3,57.3,0,128,0C198.7,0,256,57.3,256,128z};
    \fill[white] svg{M86.3,186.2H70.9V79.1h15.4v48.4V186.2z}
                 svg{M108.9,79.1h41.6c39.6,0,57,28.3,57,53.6c0,27.5-21.5,53.6-56.8,53.6h-41.8V79.1z M124.3,172.4h24.5c34.9,0,42.9-26.5,42.9-39.7c0-21.5-13.7-39.7-43.7-39.7h-23.7V172.4z}
                 svg{M88.7,56.8c0,5.5-4.5,10.1-10.1,10.1c-5.6,0-10.1-4.6-10.1-10.1c0-5.6,4.5-10.1,10.1-10.1C84.2,46.7,88.7,51.3,88.7,56.8z};
  }
}
\newcommand\orcidicon[1]{\href{https://orcid.org/#1}{\mbox{\scalerel*{
\begin{tikzpicture}[yscale=-1,transform shape]
\pic{orcidlogo};
\end{tikzpicture}
}{|}}}}

\usepackage{hyperref} 
\usepackage{graphicx}
\usepackage{amsmath}
\usepackage{amssymb}
\usepackage{multirow}
\usepackage{amsthm}
\usepackage{booktabs}
\usepackage{makecell}  
\usepackage{scalerel} 
\usepackage{diagbox} 
\usepackage[ruled,vlined]{algorithm2e}
\usepackage{subfigure}
\usepackage[normalem]{ulem}

\usepackage[export]{adjustbox} 
\usepackage{wrapfig} 
\usepackage{placeins}  
\usepackage{needspace}
\usepackage{url}
\usepackage{siunitx}
\usepackage[normalem]{ulem}

\begin{document}

\title{Regular Polytope Networks}

\author{Federico~Pernici,
        Matteo~Bruni,
        Claudio~Baecchi
        and~Alberto~Del~Bimbo,~\IEEEmembership{Member,~IEEE,}
\thanks{MICC, Media Integration and Communication Center, University of Florence, Dipartimento di Ingegneria dell'Informazione Firenze, Italy.}
}


\maketitle

\begin{abstract}
Neural networks are widely used as a model for classification in a large variety of tasks. Typically, a learnable transformation (i.e. the classifier) is placed at the end of such models returning a value for each class used for classification. This transformation plays an important role in determining how the generated features change during the learning process.
In this work, we argue that this transformation not only can be fixed (i.e. set as non-trainable) with no loss of accuracy and with a reduction in memory usage, but it can also be used to learn stationary and maximally separated embeddings.
We show that the stationarity of the embedding and its maximal separated representation can be theoretically justified by setting the weights of the fixed classifier to values taken from the coordinate vertices of the three regular polytopes available in $\mathbb{R}^d$, namely: the $d$-Simplex, the $d$-Cube and the $d$-Orthoplex. These regular polytopes have the maximal amount of symmetry that can be exploited to generate stationary features angularly centered around their corresponding fixed weights.
Our approach improves and broadens the concept of a fixed classifier, recently proposed in \cite{hoffer2018fix}, to a larger class of fixed classifier models. 
Experimental results confirm the theoretical analysis, the generalization capability, the faster convergence and the improved performance of the proposed method. Code will be publicly available.
\end{abstract}


\section{Introduction}
\IEEEPARstart{D}{eep} Convolutional Neural Networks (DCNNs) have achieved state-of-the-art performance on a variety of tasks \cite{lecun2015deep, schmidhuber2015deep} and have revolutionized Computer Vision in both classification \cite{Zoph_2018_CVPR, touvron2019FixRes} and representation \cite{Cao18, deng2019arcface}. 
In DCNNs, both representation and classification are typically jointly learned in a single network. 
The classification layer placed at the end of such models transforms the $d$-dimension of the network internal feature representation to the $K$-dimension of the output class probabilities. Despite the large number of trainable parameters that this layer adds to the model (i.e. $d \times K$), it has been verified that its removal only causes a slight increase
in error \cite{zeiler2014visualizing}. 
Moreover, the most recent architectures tend to avoid the use of fully connected layers \cite{lin2013network} \cite{szegedy2015going} \cite{he2016deep}. 
It is also well known that DCNNs can be trained to perform metric learning without the explicit use of a classification layer \cite{chopra2005learning} \cite{schroff2015facenet} \cite{hoffer2015deep}.
In particular, it has been shown that excluding from learning the parameters of the classification layer causes little or no decline in performance while allowing a reduction in the number of trainable parameters \cite{hoffer2018fix}. 
Fixed classifiers also have an important role in the theoretical convergence analysis of training models with batch-norm \cite{Ioffe17}. Very recently it has been shown that DCNNs with a fixed classifier and batch-norm in each layer establish a principle of equivalence between different learning rate schedules \cite{li2019exponential}.

All these works seem to suggest that the final fully connected layer used for classification is somewhat redundant and does not have a primary role in learning and generalization. In this paper we show that a special set of fixed classification layers \emph{has a key role in modeling the internal feature representation} of DCNNs, while ensuring little or no loss in classification accuracy and a significant reduction in memory usage.

In DCNNs the internal feature representation for an input sample is the feature vector $\mathbf{f}$ generated by the penultimate layer, while the last layer (i.e. the classifier) outputs score values according to the inner product as: 
\begin{equation}
z_i = \mathbf{w}_i^\top \cdot \mathbf{f}
\label{eq_logit}
\end{equation}
for each class $i$, where $\mathbf{w}_i$ is the weight vector of the classifier for the class $i$. To evaluate the loss, the scores are further normalized into probabilities via the softmax function \cite{goodfellow2016deep}.
\begin{figure}[t]
\includegraphics[width=0.99\columnwidth]{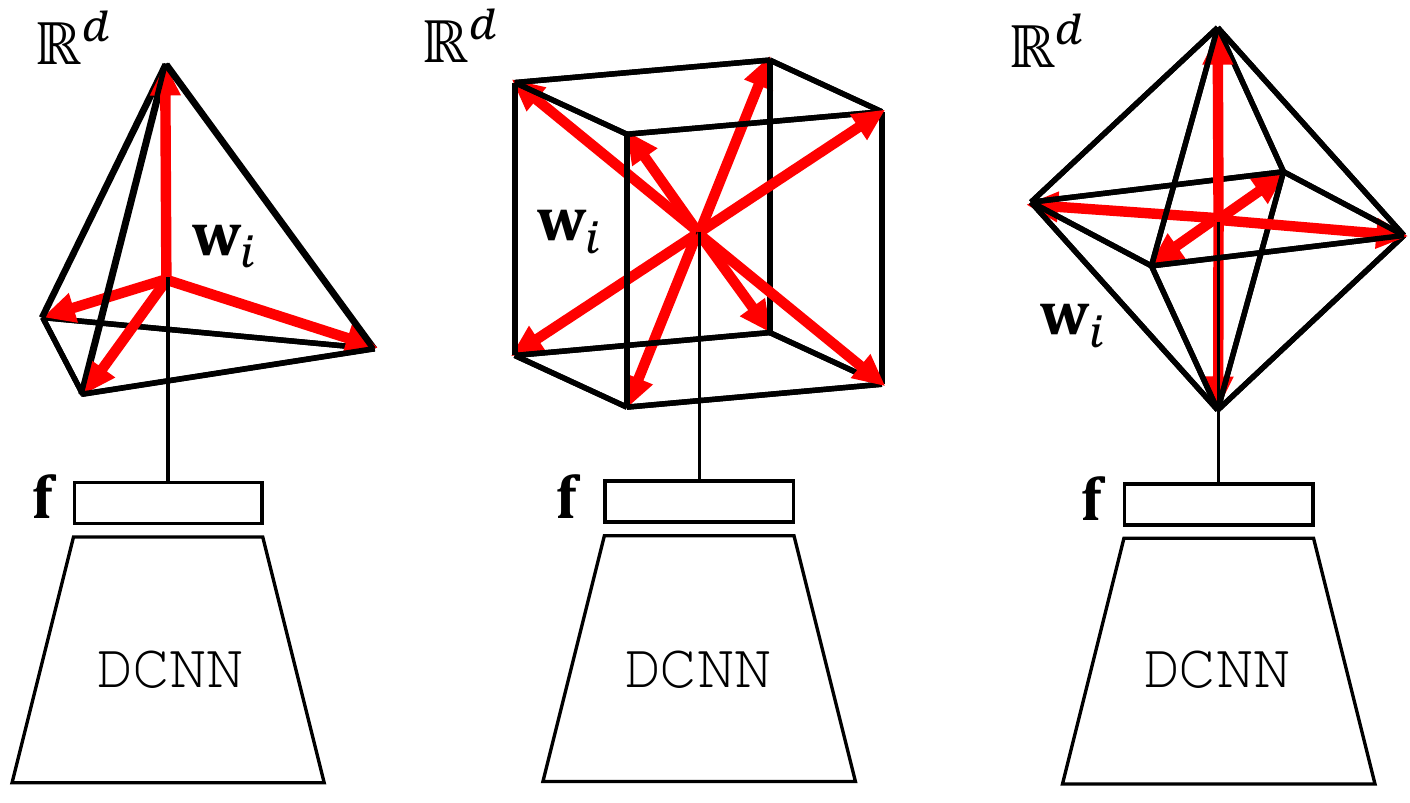}
\caption{ Regular Polytope Networks (RePoNet). 
The fixed classifiers derived from the three regular polytopes available in $\mathbb{R}^d$ with $d \geq 5$ are shown. 
From left: the $d$-Simplex, the $d$-Cube and the $d$-Orthoplex fixed classifier. The trainable parameters $\mathbf{w}_i$ of the classifier are replaced with fixed values taken from the coordinate vertices of a regular polytope (shown in red).
}
\label{fig_IntroFig}
\end{figure}
Since the values of $z_i$ can be also expressed as $z_i = \mathbf{w}^\top_i \cdot \mathbf{f} = ||\mathbf{w}_i|| \: ||\mathbf{f}|| \cos(\theta)$, where $\theta$ is the angle between $\mathbf{w}_i$ and $\mathbf{f}$, the score for the correct label with respect to the other labels is obtained by optimizing the length of the vectors $||\mathbf{w}_i||$, $||\mathbf{f}||$ and the angle $\theta$ they are forming. 
This simple formulation of the final classifier provides the intuitive explanation of how feature vector directions and weight vector directions align simultaneously with each other at training time so that their average angle is made as small as possible.
\begin{figure*}[htbp]
\hspace{-0.56cm}
\centering
\subfigure[]{\label{fig:a}\includegraphics[height=0.178\linewidth,valign=t]{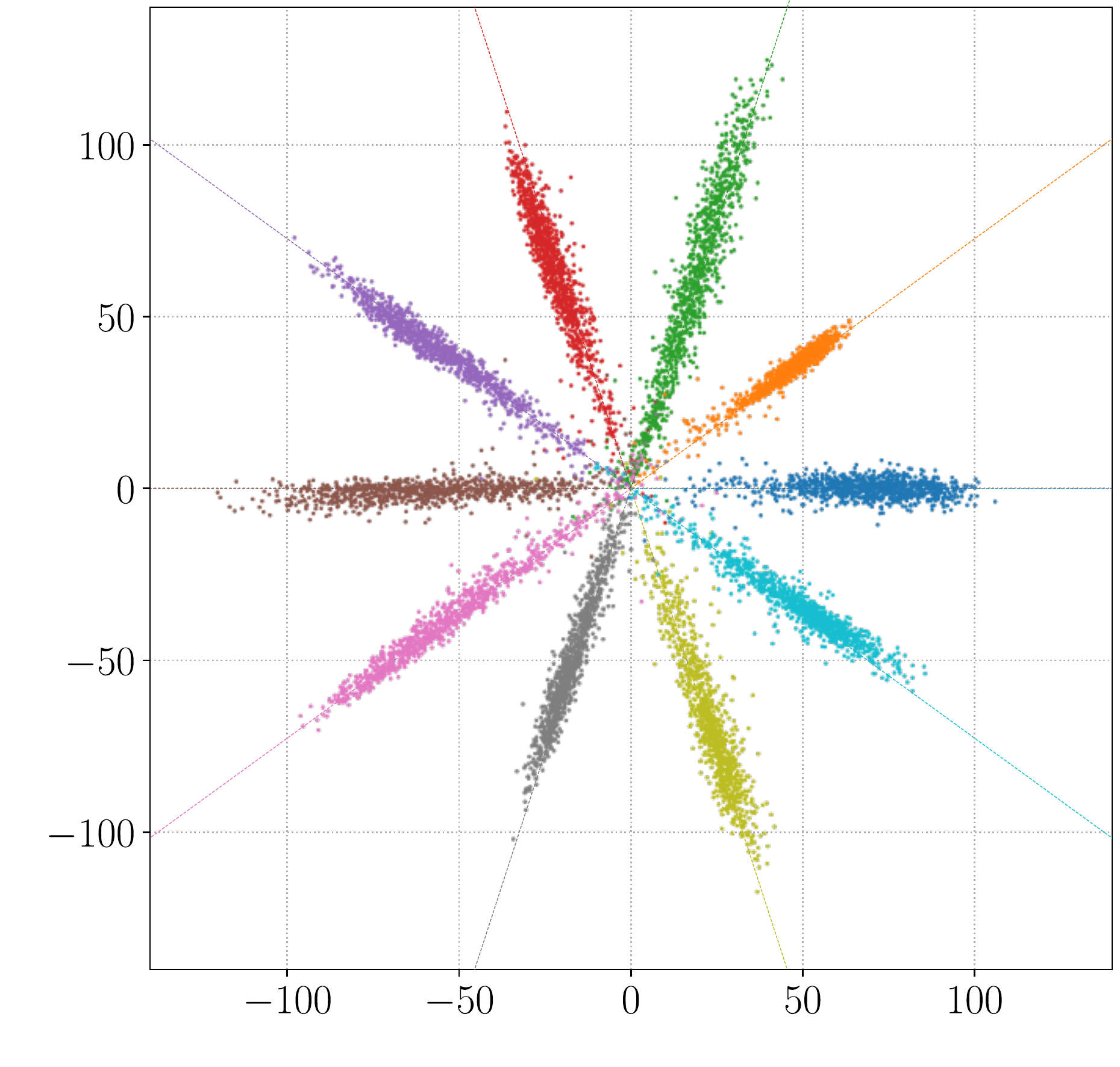}}
\subfigure[]{\label{fig:b}\includegraphics[height=0.178\linewidth,valign=t]{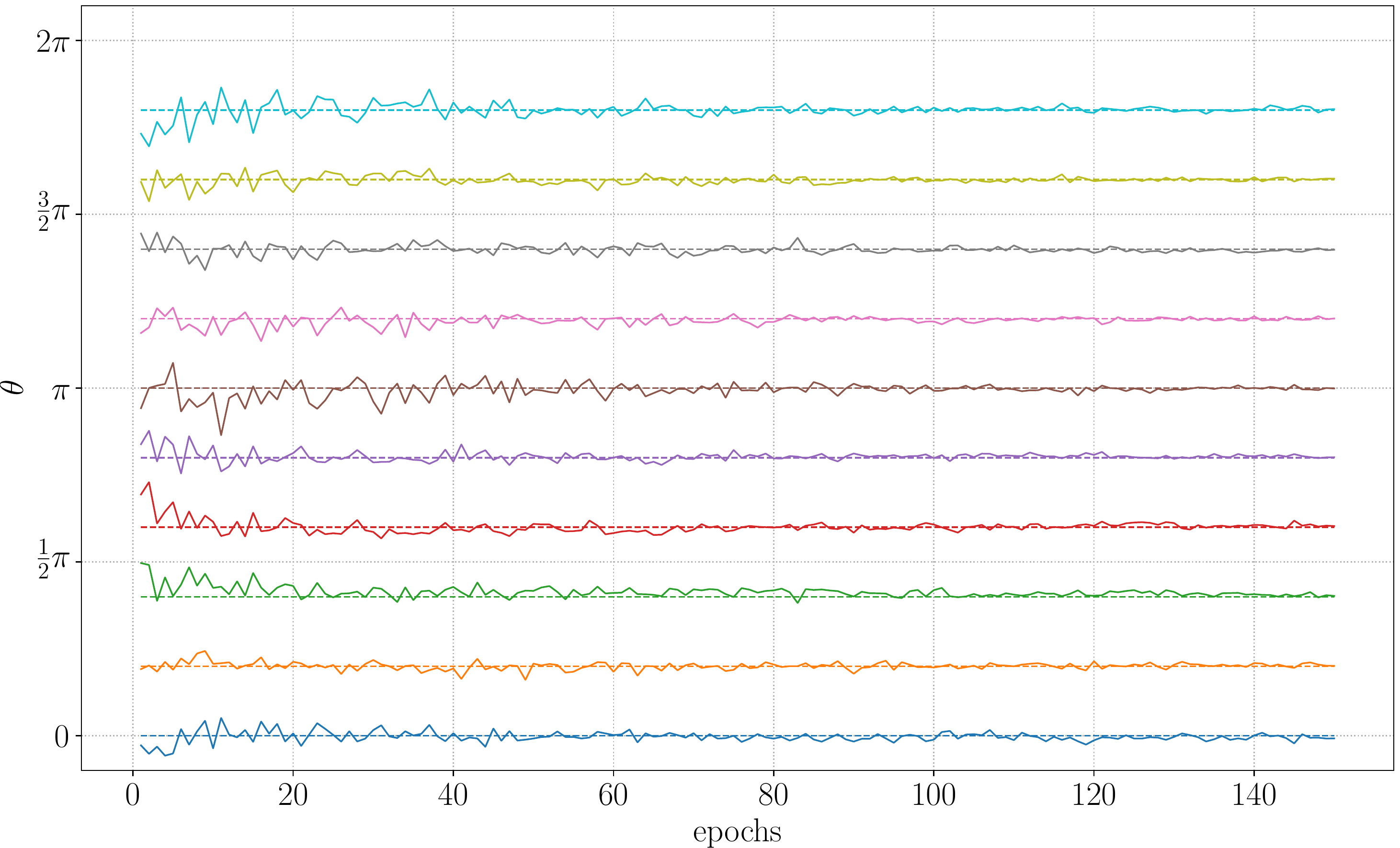}}\hspace{1pt}
\subfigure[]{\label{fig:c}\includegraphics[height=0.178\textwidth,valign=t]{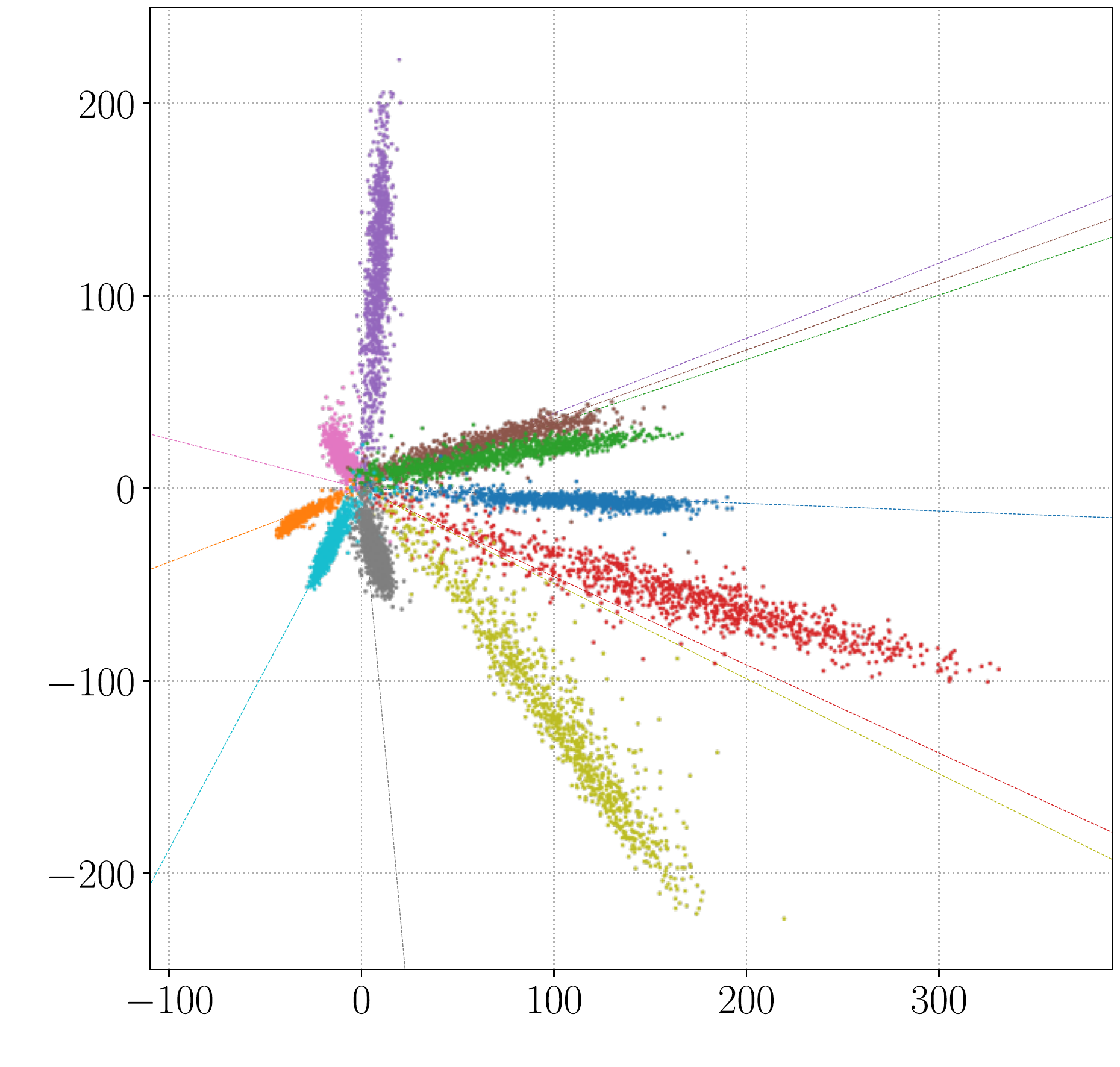}}%
\subfigure[]{\label{fig:d}\includegraphics[height=0.178\textwidth,valign=t]{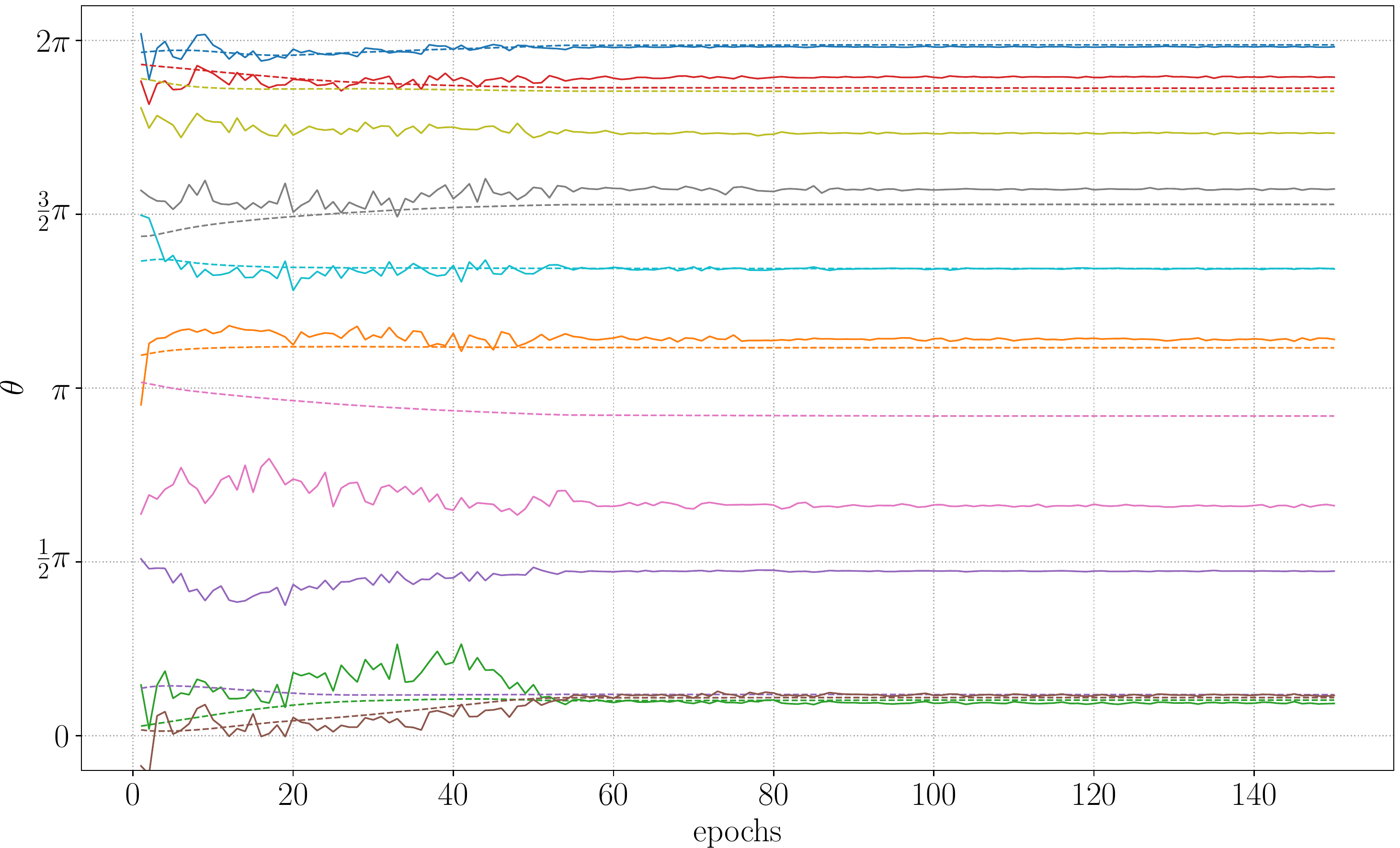}}\hspace{1pt}%
\raisebox{-0.6pt}{\includegraphics[height=0.1609\textwidth,width=0.047\textwidth,valign=t]{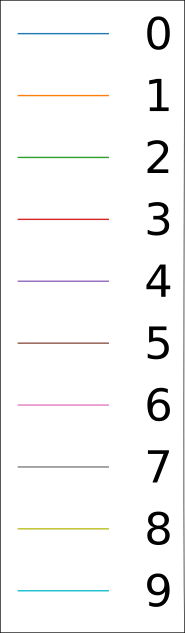}}
\caption{Feature learning on the MNIST dataset in a 2D embedding space. Fig.~(a) and Fig.~(c) show the 2D features learned by RePoNet and by a standard trainable classifier respectively.
Fig.~(b) and Fig.~(d) show the training evolution of the classifier weights (dashed) and their corresponding class feature means (solid) respectively. Both are expressed according to their angles. Although the two methods achieve the same classification accuracy, features in the proposed  method are both stationary and maximally separated.
}
\label{fig:intro_fig_mnist}
\end{figure*}
If the parameters $\mathbf{w}_i$ of the classifier in Eq.~\ref{eq_logit} are fixed  (i.e. set as non-trainable), \emph{only the feature vector directions} can align toward the classifier weight vector directions and not the opposite. Therefore, weights can be regarded as fixed angular references to which features align.

According to this, we obtain a precise result on the spatio-temporal statistical properties of the generated features during the learning phase. Supported by the empirical evidence in \cite{hoffer2018fix} we show that not only the final classifier of a DCNN can be set as non-trainable with no loss of accuracy and with a significant reduction in memory usage, but that an appropriate set of values assigned to its weights allows learning a maximally separated and strictly stationary embedding while training. That is, the features generated by the Stochastic Gradient Descent (SGD) optimization have constant mean and are angularly centered around their corresponding fixed class weights. 
Constant known mean implies that features cannot have non-constant trends while learning.
Maximally separated features and their stationarity are obtained by setting the classifier weights according to values following a highly symmetrical configuration in the embedding space. 

DCNN models with trainable classifiers are typically convergent
and therefore, after a sufficient learning time has elapsed, some form of stationarity in the learned features can still be achieved. However, until that time, it is not possible to  know where the features will be projected by the learned model in the embedding space. 
An advantage of the approach proposed in this paper is that it allows to define (and therefore to know in advance) where the features will be projected before starting the learning process.

Our result can be understood by looking at the basic functionality of the final classifier in a DCNN. The main role of a trainable classifier is to dynamically adjust the decision boundaries to learn class feature representations. When the classifier is set as non-trainable this dynamic adjustment capability is no longer available and it is automatically demanded to all the previous layers. Specifically, the work \cite{hoffer2018fix} reports  empirical evidence that the expressive power of DCNN models is large enough to account for the missing dynamic adjustment capability of the  classifier.
We provide more systematic empirical evidence confirming and broadening the general validity of DCNNs with fixed classifiers (Sec.~\ref{sec_Exchangeability_Assumption}).

We show that our approach can be theoretically justified and easily implemented by setting the classifier weights to values taken from the coordinate vertices of a regular polytope in the embedding space.
Regular polytopes are the generalization in any number of dimensions of regular polygons and regular polyhedra (i.e. Platonic Solids). Although there are infinite regular polygons in $\mathbb{R}^2$ and 5 regular polyhedra in $\mathbb{R}^3$, there are only three regular polytopes in $\mathbb{R}^d$ with $d \geq 5$, namely the $d$-Simplex, the $d$-Cube and the $d$-Orthoplex. Having different symmetry, geometry and topology, each regular polytope will reflect its properties into the classifier and the embedding space which it defines. Fig.~\ref{fig_IntroFig} illustrates the three basic architectures defined by the proposed approach termed Regular Polytope Networks (RePoNet). Fig.~\ref{fig:intro_fig_mnist} provides a first glance at our main result in a 2D embedding space. Specifically, the main evidence from Fig.~\ref{fig:a} and \ref{fig:b} is that the features learned by RePoNet remain aligned with their corresponding fixed weights and maximally exploit the available representation space directly from the beginning of the training phase.

We apply our method to multiple vision datasets showing that it is possible to generate stationary and maximally separated features without affecting the generalization performance of DCNN models and with a significant reduction in GPU memory usage at training time. A preliminary exploration of this work was presented in \cite{Pernici_2019_CVPR_Workshops, pernici2020class}.

\section{Related Work}

\textbf{Fixed Classifier}. 
Empirical evidence shows that convolutional neural networks with a fixed classification layer (i.e. not subject to learning) initialized by random numbers does not worsen the performance on the CIFAR-10 dataset \cite{hardt2016identity}. A recent paper \cite{hoffer2018fix} explores in more detail the idea of excluding from learning the parameters $\mathbf{w}_i$ in Eq.\ref{eq_logit}. The work shows that a fixed classifier causes little or no reduction in classification performance for common datasets while allowing a significant reduction in trainable parameters, especially when the number of classes is large. Setting the last layer as not trainable also reduces the computational complexity for training as well as the communication cost in distributed learning. The  paper in question sets the classifier with the coordinate vertices of orthogonal vectors taken from the columns of the Hadamard\footnote{The Hadamard matrix is a square matrix whose entries are either $+1$ or $-1$ and whose rows are mutually orthogonal.} matrix and does not investigate on the internal feature representation.
A major limitation of this method is that, when the number of classes is higher than the dimension of the feature space, it is not possible to have mutually orthogonal columns. As a consequence some of the classes are constrained to lie in a common subspace which causes a reduction in classification performance. 
In our solution, we improve and generalize this work by finding a novel set of unique directions overcoming the limitations of the Hadamard matrix. 

The work \cite{sablayrolles2018spreading} trains a neural network according to the triplet loss with a set of fixed vertices on a hyper-sphere (i.e. a sphere lattice). The work aims at learning a function that maps real-valued vectors to a uniform distribution over a $d$-dimensional sphere. 

As shown in \cite{li2019exponential}, fixed classifiers are also related to BatchNorm \cite{Ioffe17} and learning rate schedules. 
BatchNorm parametrizes the weights of a layer to “normalize” its activations (i.e. the features), but typically it is not applied to normalize the outputs of the classifier (i.e. logits). 
They show that, with BatchNorm layers and a fixed classifier layer, training with $L2$ regularization is equivalent to training with an exponentially increasing learning rate.

\textbf{Softmax Angular Optimization}.
As originally described in \cite{liu2016large}, under softmax loss\footnote{The combination of cross-entropy loss and the softmax function at the last fully connected layer.} the class prediction is \emph{largely} determined by the angular similarity since softmax loss can be factorized, as shown in Eq.~\ref{eq_logit}, into an amplitude component and an angular component.
Several papers have followed this intuition and proposed to train DCNNs by direct angle optimization \cite{ranjan2017l2,Liu2017CVPR,Liu_2018_CVPR,wang2017normface}. The angle encodes the required discriminative information for class recognition. The wider the angles the better the classes are separated from each other and, accordingly, their representation is more discriminative.  The common idea of these works is to constrain the features and/or the classifier to be unit normalized. 
The works \cite{liu_2017_coco_v1}, \cite{hasnat2017mises} and \cite{wang2017normface} normalize both features and the classifier weights thus obtaining an exact optimization of the angle in Eq.~\ref{eq_logit}. 
With weight normalization only, label prediction is largely determined by the angular similarity \cite{liu2016large}   \cite{Liu2017CVPR}. This is not only because Eq.~\ref{eq_logit} can be e factorized into amplitude and angular
component, but also because decision boundaries between adjacent classes are determined by their angular bisectors.

Differently from weight normalization, feature normalization cannot directly perform angle optimization but encourages intra-class compactness of learned features \cite{ranjan2017l2}.  
Specifically, \cite{ranjan2017l2} also proposes adding a multiplicative scale parameter after feature normalization based on the property that increasing the norm of samples can decrease the softmax loss \cite{wang2017normface,yuan2017feature}.
Although with a different goal than learning discriminative features, the work \cite{hoffer2018fix}, in addition to fixing the classifier, normalizes both the weights and the features and applies the multiplicative scale parameter.

In agreement with \cite{wang2018additive,hoffer2018fix,ranjan2017l2} and \cite{wang2017normface} we found that applying the feature normalization and the multiplicative scale parameter 
makes optimization hard with general datasets, having a significant dependence on image quality. According to this, we follow the work \cite{Liu2017CVPR} that normalizes the classifier weights. 
Normalizing classifier weights typically also includes setting the classifier biases to zero. As discussed in \cite{wang2017normface} and in \cite{yuan2017feature} this encourages well-separated features to have bigger magnitudes. This avoids features collapsing into the origin, making angles between the weights and features a reliable metric for classification.

As conjectured in \cite{wang2017normface}, if all classes are well-separated, weight normalization  will roughly correspond to computing the mean of features in each class. 
The maximal and fixed separation proposed in this paper further strengthens the conjecture, producing features more centered around their fixed weights as the training process progresses.

Another close related work to ours is \cite{virtualclass} in which separability of learned features is improved by injecting a single dynamic virtual negative class into the original softmax. A virtual class is a class that is active in the classifier but has no data available from which to learn.
Injecting the virtual class enlarges the inter-class margin and compresses intra-class distribution by strengthening the decision boundary constraint. 
In our case, we can profitably exploit virtual classes when the number of classes of the fixed classifier does not match the number of vertices of a regular polytope.

While all the above works impose large angular distances between the classes, they provide solutions to enforce such constraint in a local manner without considering global inter-class separability and intra-class compactness. For this purpose, very recently the works \cite{HypersphericalEnergy2018}, \cite{Zhao_2019_CVPR} and \cite{Duan_2019_CVPR} add a regularization loss to specifically force the classifier weights to be far from each other in a global manner. These works draw inspiration from a well-known problem in physics -- the Thomson problem \cite{thomson1904xxiv}, where given $K$ charges confined to the surface of a sphere, one seeks to find an arrangement of the charges which minimizes the total electrostatic energy. Electrostatic force repels charges each other inversely proportional to their mutual distance. In \cite{HypersphericalEnergy2018}, \cite{Zhao_2019_CVPR} and \cite{Duan_2019_CVPR} global equiangular features are obtained by adding to the standard categorical cross-entropy loss a further loss inspired by the Thomson problem.

In our research, we follow a similar principle for global separability. We consider that minimal energies are often concomitant with special geometric configurations of charges that recall the geometry of Platonic Solids in high dimensional spaces \cite{batle2016generalized}.
We have reported a few preliminary and  qualitative results of using Regular Polytope Networks for compact feature learning in \cite{Pernici_2019_CVPR_Workshops}, where we have demonstrated that the angular parameter of the margin loss of \cite{deng2019arcface} can be analytically determined to maximize feature compactness.

\section{Main Contributions:}
Our technical contributions can be summarized as follows:
\begin{enumerate}
\item We generalize the concept of fixed classifiers and show they can generate stationary and maximally separated features at training time with no loss of performance and in many cases with slightly improved performance.
\item We performed extensive evaluations across a range of datasets and modern CNN architectures reaching state-of-the-art performance. We observed faster speed of convergence and a significant reduction in model parameters.
\item We further provide a formal characterization of the class decision boundaries according to the dual relationship between regular polytopes and statistically verify the validity of our method on random permutations of the labels.
\end{enumerate}

\section{Regular Polytopes and Maximally Separated Stationary Embeddings}
We are basically concerned with the following question: \emph{How should the non-trainable weights of the classifier be distributed in the embedding space such that they generate stationary and maximally separated features?}

Let $\mathbb{X} = \{(x_i, y_i)\}_{i=1}^{N}$ be the training set containing $N$ samples, where $x_i$ is the raw input to the DCNN and $y_{i} \in \{1,2,\cdots,K\}$ is the label of the class that supervises the output of the DCNN. Then, the cross entropy loss can be written as:
\begin{equation}
\mathcal{L}=-\frac{1}{N}\sum_{i=1}^{N} \log\Bigg( \frac {\exp({\mathbf{w}_{y_i}^{\top}\mathbf{f}_{i}+\mathbf{b}_{y_i}})} {\sum_{j=1}^{K}\exp({\mathbf{w}_{j}^{\top}\mathbf{f}_{i} + \mathbf{b}_{j}})} \Bigg), 
\label{softmax_loss}
\end{equation}
where $\mathbf{W}=\{ \mathbf{w}_j \}_{j=1}^{K}$ are the classifier weight vectors for the $K$ classes.
Following the discussion in \cite{Liu2017CVPR} we normalize the weights and zero the biases ($\hat{\mathbf{w}}_j = \frac{\mathbf{w}_j}{||\mathbf{w}_j||}$, $\mathbf{b}_j=0$) to directly optimize angles, enabling the network to learn angularly distributed features.
Angles therefore encode the required discriminative information for class recognition and the wider they are, the better the classes are represented. As a consequence, the representation in this case is maximally separated when features are distributed at \emph{equal angles} maximizing the available space. 

If we further consider the feature vector parametrized by its unit vector as $\mathbf{f}_i=\kappa_i \, \hat{\mathbf{f}}_i$ where $\kappa_i=||\mathbf{f}_i||$ and $\hat{\mathbf{f}}_i = \frac{\mathbf{f}_i}{||\mathbf{f}_i||}$, then Eq.\ref{softmax_loss} can be rewritten as: 
\begin{equation}
\mathcal{L}=-\frac{1}{N}\sum_{i=1}^{N}\log\Bigg( \frac {\exp ( { \kappa_i \hat{\mathbf{w}}_{y_i}^{\top}\hat{\mathbf{f}}_i })} {\sum_{j=1}^{K}\exp({ \kappa_i \hat{\mathbf{w}}_{j}^{\top}\hat{\mathbf{f}}_i })} \Bigg)  
\label{softmax_loss_von}
\end{equation}
The equation above can be interpreted as if $N$ realizations from a set of $K$ von Mises-Fisher distributions with different concentration parameters $\kappa_i$ are passed through the softmax function. The probability density function of the von Mises-Fisher distribution for the random $d$-dimensional unit vector $\hat{\mathbf{f}}$ is given by:  $ P(\hat{\mathbf{f}} ;\hat{\mathbf{w}},\kappa ) \propto \exp  \big ( {\kappa \hat{\mathbf{w}}^{\top} \hat{\mathbf{f}} } \big ) $ where $ \kappa \geq 0$. Under this parameterization $\hat{\mathbf{w}}$ is the mean direction on the hypersphere and $\kappa$ is the concentration parameter. The greater the value of $\kappa$ the higher the concentration of the distribution around the mean direction $\hat{\mathbf{w}}$. The distribution is unimodal for $\kappa>0$ and is uniform on the sphere for $\kappa=0$. 

As with this formulation each weight vector is the mean direction of its associated features on the hypersphere, equiangular features maximizing the available space can be obtained by arranging accordingly their corresponding weight vectors around the origin. This problem is equivalent to distributing points uniformly on the sphere and is a well-known geometric problem, called Tammes problem  \cite{tammes1930origin} which is a generalization of the physic problem firstly addressed by Thomson \cite{thomson1904xxiv}. 
In 2D the problem is that of placing $K$ points on a circle so that they are as far as possible from each other. In this case the optimal solution is that of placing the points at the vertices of a regular $K$-sided polygon. 
The 3D analogous of regular polygons are Platonic Solids. However, the five Platonic solids are not always the unique solutions of the Thomson problem. In fact, only the tetrahedron, octahedron and the icosahedron are the unique solutions for $K = 4$, $6$ and $12$ respectively. For $K = 8$: the cube is not optimal in the sense of the Thomson problem. This means that the energy stabilizes at a minimum in configurations that are not symmetric from a geometric point of view. The unique solution in this case is provided by the vertices of an irregular polytope \cite{bagchi1997stay}.  

The non geometric symmetry between the locations causes the global charge to be different from zero. Therefore in general, when the number of charges is arbitrary, their position on the sphere cannot reach a configuration for which the global charge vanishes to zero. A similar argument holds in higher dimensions for the so called generalized Thomson problem \cite{batle2016generalized}. According to this, we argue that, \emph{the geometric limit to obtain a zero global charge in the generalized Thomson problem is equivalent to the impossibility to learn  maximally separated features for an arbitrary number of classes}. 

However, since classification is not constrained in a specific dimension as in the case of charges, our approach addresses this issue by \emph{selecting the appropriate dimension of the embedding space so as to have access to symmetrical fixed classifiers directly from regular polytopes}.  
In dimensions five and higher, there are only three ways to do that (See Tab.~\ref{tab_polytopes}) and they involve the symmetry properties of the three well known regular polytopes available in any high dimensional spaces \cite{coxeter1963regular}. These three special classes exist in every dimension and are: the $d$-Simplex, the $d$-Cube and the $d$-Orthoplex. 
In the next paragraphs the three fixed classifiers derived from them are presented.
\begin{table}[t]
\centering
\caption{Number of regular Polytopes as dimension $d$ increases.}
\begin{tabular}{@{}l|lllll@{}}
\toprule
\toprule
Dimension $d$ & $1$ & $2$ & $3$ & $4$ & $\geq 5$ \\ \midrule
Number of Regular Polytopes & $1$ & $\infty$ & $5$ & $6$ & $\mathbf{3}$ \\ 
\bottomrule
\bottomrule
\end{tabular}
\label{tab_polytopes}
\end{table}

\textbf{The $d$-Simplex Fixed Classifier.}
In geometry, a simplex is a generalization of the notion of a triangle or tetrahedron to arbitrary dimensions. Specifically, a $d$-Simplex is a $d$-dimensional polytope which is the convex hull of its $d + 1$ vertices. A regular $d$-Simplex may be constructed from a regular $(d-1)$-Simplex by connecting a new vertex to all original vertices by the common edge length. According to this, the weights for this classifier can be computed as: 
\begin{equation}
\mathbf{W}_\mathcal{S}=\Big \{e_1,e_2,\dots,e_{d-1}, \alpha \sum_{i=1}^{d-1} e_i \Big \}
\nonumber
\end{equation}
where $\alpha=\frac{1-\sqrt{d+1}}{d}$ and $e_i$ with $i \in \{1,2, \dots, d-1\}$ denotes the standard basis in $\mathbb{R}^{d-1}$. The final weights will be shifted about the centroid and normalized.
The $d$-Simplex fixed classifier defined in an embedding space of dimension $d$ can accommodate a number of classes equal to its number of vertices:
\begin{equation}
K=d+1.
\label{eq_ksimplex}
\end{equation}
This classifier has the largest number of classes that can be embedded in $\mathbb{R}^d$ such that their corresponding class features are equidistant from each other. It can be shown (see appendix) that the angle subtended between \emph{any} pair of weights is equal to: 
\begin{equation}
\theta_{\mathbf{w}_i,\mathbf{w}_j}=\arccos\bigg(-\frac{1}{d}\bigg) \quad \forall i,j \in \{ 1, 2, \dots, K\} : i \neq j.
\label{eq_dsimplex_angle}
\end{equation}
\textbf{The $d$-Orthoplex Fixed Classifier.}
This classifier is derived from the $d$-Ortohoplex (or Cross-Polytope) regular polytope that is defined by the convex hull of points, two on each Cartesian axis of an Euclidean space, that are equidistant from the origin. The weights for this classifier can therefore be defined as:
$$\mathbf{W}_\mathcal{O} = \{ \pm e_1,\pm e_2,\dots,\pm e_d \}.$$
Since it has $2d$ vertices, the derived fixed classifier can accommodate in its embedding space of dimension $d$ a number of distinct classes equal to:
\begin{equation}
K=2d.
\label{eq_kortho}
\end{equation}
Each vertex is adjacent to other $d-1$ vertices and the angle between adjacent vertices is 
\begin{equation}
\theta_{\mathbf{w}_i,\mathbf{w}_j}= \frac{\pi}{2} \quad \forall  \, i,j  \in \{ 1, 2, \dots, K\}  : j \in C(i)
\label{eq_dortho_angle}
\end{equation}
Where each $j \in C(i)$ is an adjacent vertex and $C$ is the set of adjacent vertices defined as ${C}(i)=\{ j:(i,j) \in {E}\}$. $E$ is the set of edges of the graph ${G}=(\mathbf{W}_\mathcal{O},{E})$.
The $d$-Orthoplex is the dual polytope of the $d$-Cube and vice versa (i.e. the normals of the $d$-Orthoplex faces correspond to the the directions of the vertices of the $d$-Cube).

\textbf{The $d$-Cube Fixed Classifier.}
The $d$-Cube (or Hypercube) is the regular polytope formed by taking two congruent parallel hypercubes of dimension $(d-1)$ and joining pairs of vertices, so that the distance between them is $1$. A $d$-Cube of dimension 0 is one point.
The fixed classifier derived from the $d$-Cube
is constructed by creating a vertex for each binary number in a string of $d$ bits. Each vertex is a $d$-dimensional boolean vector with binary coordinates $-1$ or $1$. Weights are finally obtained from the normalized vertices:
\begin{equation} 
\mathbf{W}_\mathcal{C} = \left  \{ \mathbf{w} \in \mathbb {R}^{d}:  \left [-\frac{1}{\sqrt{d}},\frac{1}{\sqrt{d}} \right ] ^d \right \}.
\nonumber
\end{equation}
The $d$-Cube can accommodate a number of distinct classes equal to: 
\begin{equation}
K=2^d.
\label{eq_kcube}
\end{equation}
The vertices are connected by an edge whenever the Hamming distance of their binary numbers is one therefore forming a $d$-connected graph.
It can be shown (see appendix) that the angle between a vertex with its adjacent (i.e. connected) vertices is:
\begin{equation}
\theta_{\mathbf{w}_i,\mathbf{w}_j} = \arccos\bigg(\frac{d-2}{d}\bigg),  \forall  \, i,j  \in \{ 1, \dots, K\}  : j \in C(i)
\label{eq_dcube_angle}
\end{equation}
where $C(i)$ is the set of vertices adjacent to vertex $i$.

\begin{figure}[t]
\centering
\includegraphics[width=0.99\columnwidth]{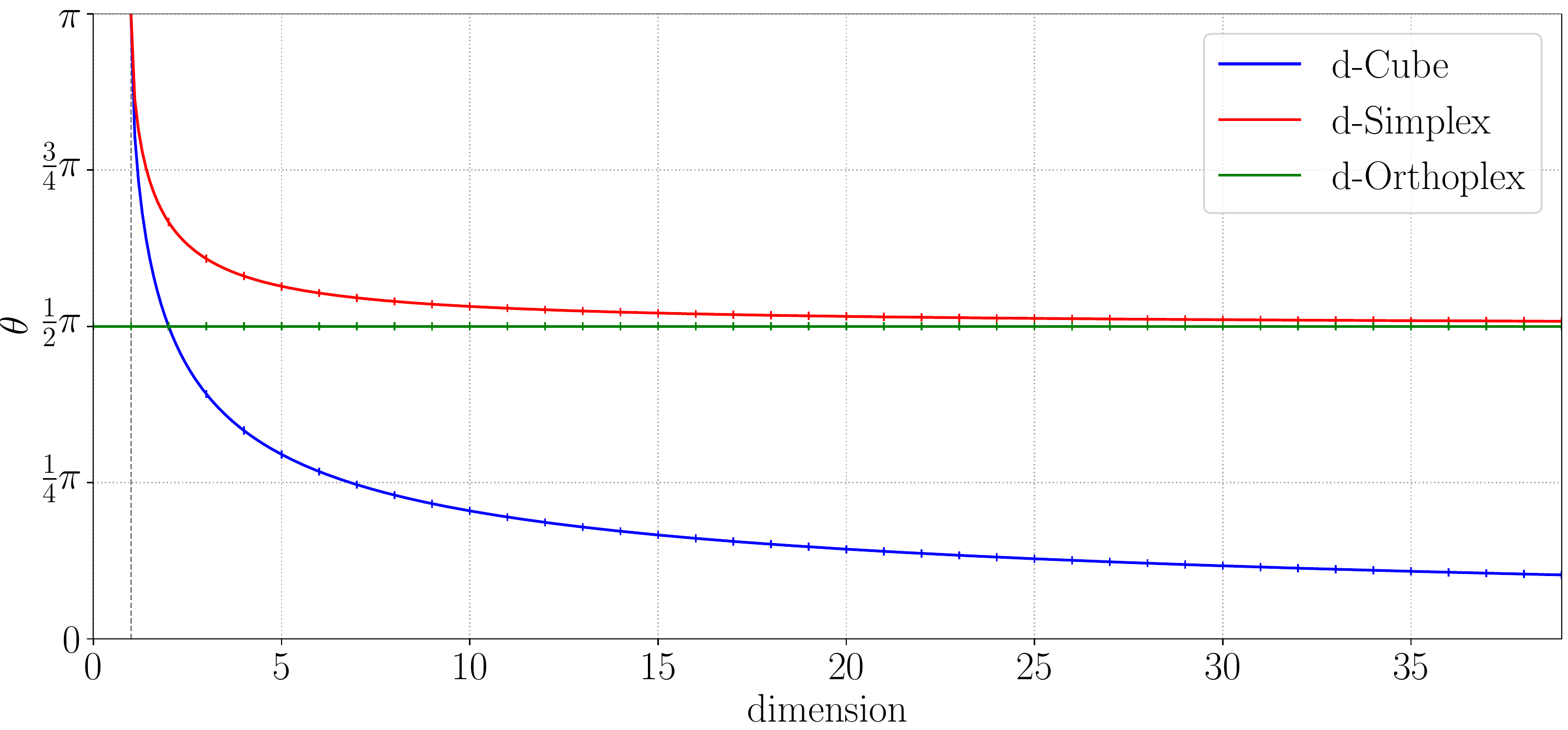}
\caption{The angular space defined by RePoNet classifiers. Curves represent the angle between a weight and its adjacent weights as the dimension of the embedding space increases. The angle between class features follows the same trend.}
\label{fig:anglefcn}
\end{figure}

Fig.~\ref{fig:anglefcn} shows the angle between a weight and its adjacent weights computed from Eqs.~\ref{eq_dsimplex_angle}, \ref{eq_dortho_angle} and \ref{eq_dcube_angle} as the dimension of the embedding space increases. Having the largest angle between the weights, the $d$-Simplex fixed classifier achieves the best inter-class separability. However, as the embedding space dimension increases, its angle tends towards $\pi/2$. Therefore, the larger the dimension is, the more similar it becomes to the $d$-Orthoplex classifier. The main difference between the two classifiers is in their neighbor connectivity. The different connectivity of the three regular polytope classifiers has a direct influence on the evaluation of the loss. In the case of the $d$-Simplex classifier, all the summed terms in the loss of Eq.~\ref{softmax_loss_von} have always comparable magnitudes in a mini batch. 

The $d$-Cube classifier has the most compact feature embedding and the angle between each weight and its $d$ neighbors decreases as the dimension increases. Due to this, it is the hardest to optimize.

\subsection{Implementation}
Given a classification problem with $K$ classes, the three RePoNet fixed classifiers can be simply instantiated by defining a non-trainable fully connected layer of dimension $d$, where $d$ is computed from Eqs.~\ref{eq_ksimplex}, \ref{eq_kortho} and \ref{eq_kcube} as summarized in Tab~\ref{tab_layer_dim}.  
\begin{table}[h]
\centering
\caption{Feature dimension $d$ as a function of the number of classes $K$.}
\begin{tabular}{@{}c|ccl@{}}
\toprule
\toprule
RePoNet  & $d$-Simplex & $d$-Cube & $d$-Orthoplex \\ \midrule
Layer dim. & $d=K-1$ & $d = \lceil \log_2(K) \rceil$ & $d = \big \lceil \frac{K}{2} \big \rceil$ \\ \bottomrule \bottomrule
\end{tabular}
\label{tab_layer_dim}
\end{table}
\FloatBarrier
In order to accommodate different CNN architectures having different convolutional activation output size (e.g., from the 2048 size of the ResNet50 to the feature size of 10 of the fixed $d$-Cube classifier with the 1000 classes of ImageNet), a middle ``junction'' linear layer (without ReLu) is required.

\begin{figure}[t]
    \centering
    \includegraphics[width=0.75\columnwidth]{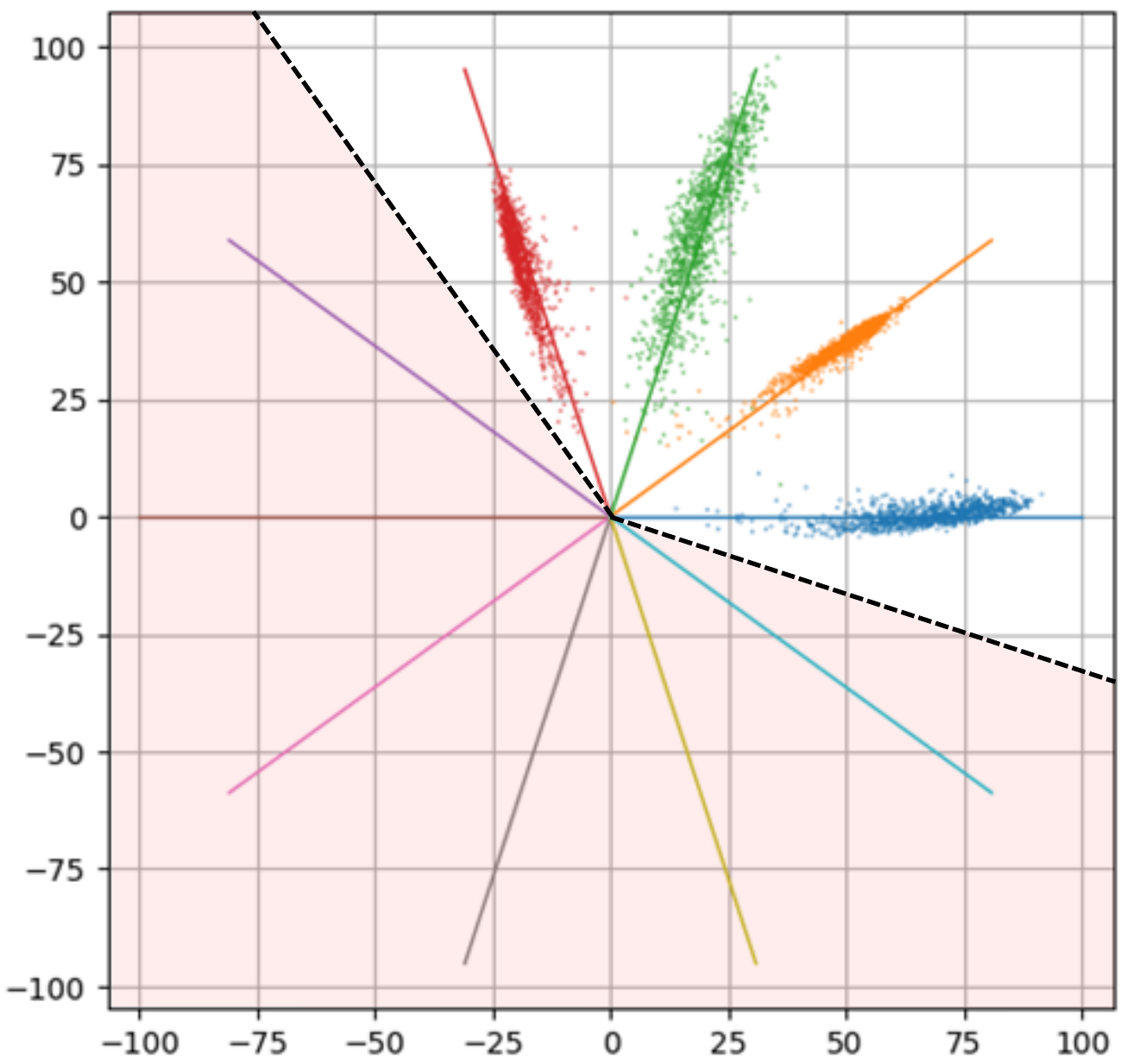}
    \caption{Learning with unassigned classes in a 2D embedding space. 
    The features of the first four digits of the MNIST dataset are learned using a 10-sided regular polygon in which six of the classes are virtual. Unassigned classes (colored lines inside the shaded region) force a large angular margin region (shaded region) from which features are pushed out.
    }
    \label{virtual_classes}
\end{figure}

\subsection{Exceeding Vertices as Virtual Negative Classes} 
Except for the $d$-Simplex that allows to assign all its vertices for a given number of classes $K$, for both the $d$-Cube and the $d$-Orthoplex classifiers some of the vertices may be in excess for a given number of classes. As implied by Eq.~\ref{eq_kortho}, in the case of the $d$-Orthoplex one vertex remains unassigned when the number of classes $K$ is odd. In the case of the $d$-Cube classifier, due to the exponential dependency in Eq.~\ref{eq_kcube}, a large number of vertices may remain not assigned. For example, assuming $K=100$ the $d$-Cube fixed classifier  has $128$ vertices (see Tab.~\ref{tab_layer_dim}) and $28$ of them are not assigned to any class.  
As shown in \cite{virtualclass}, unassigned classes act as virtual negative classes forcing a margin around the unassigned weights without affecting the correctness of softmax based cross entropy optimization. 
Virtual negative classes do not change substantially the objective function of Eq.~\ref{softmax_loss_von} that can be rewritten as:
\begin{equation}
\resizebox{0.88\hsize}{!}{$
{\displaystyle
    \mathcal{L}=-\frac{1}{N}\sum_{i=1}^{N}\log\Bigg( \frac {\exp ( { \kappa_i \hat{\mathbf{w}}_{y_i}^{\top}\hat{\mathbf{f}}_i })} {\sum_{j=1}^{K}\exp({ \kappa_i \hat{\mathbf{w}}_{j}^{\top}\hat{\mathbf{f}}_i }) + \sum_{j=K+1}^{K_V}\exp({ \kappa_i \hat{\mathbf{w}}_{j}^{\top}\hat{\mathbf{f}}_i })} \Bigg)
}
$}
\label{softmax_loss_virtual}
\end{equation}
where $K_{V}$ is the number of virtual classes (i.e. the exceeding polytope vertices).
Fig.~\ref{virtual_classes} illustrates an example similar to Fig.~\ref{fig:intro_fig_mnist}(a) in which a 10-sided polygon fixed classifier is learned to classify the first four digits of the MNIST dataset ($0,1,2$ and $3$). The remaining six ``empty slots'' of the classifier are not assigned to any class data and therefore the classifier acts as a virtual negative classifier forcing a large margin (the shaded region) around the virtual class weights (colored lines).
This result generalizes the proposed method to any arbitrary number of classes. 

\begin{figure}[t]
\centering
\includegraphics[width=0.75\columnwidth]{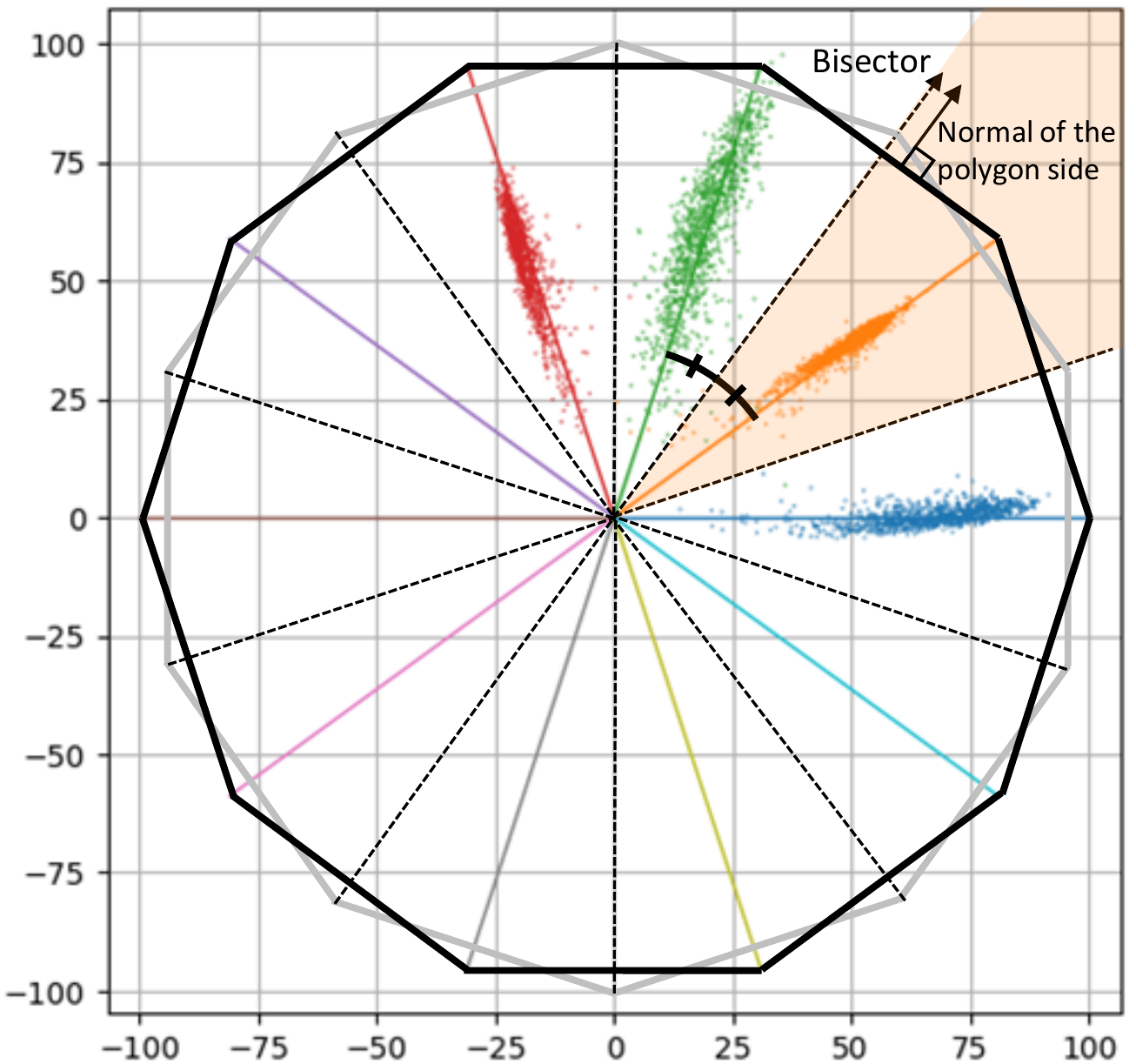}
\caption{ The intuition behind the decision boundaries in RePoNet ($10$-sided polygon).
The bisector directions (dotted lines), represent the class decision boundaries. They have the same direction of the normal of the corresponding polygon side (only one shown for clarity).
The decision boundaries form a regular polygon that is related with the classifier $10$-sided polygon according to duality. 
For clarity, only one class region is highlighted (shaded region).}
\label{polygonduality}
\end{figure}

\begin{figure}[t]
\centering
\subfigure[]{
    \includegraphics[width=0.79\columnwidth]{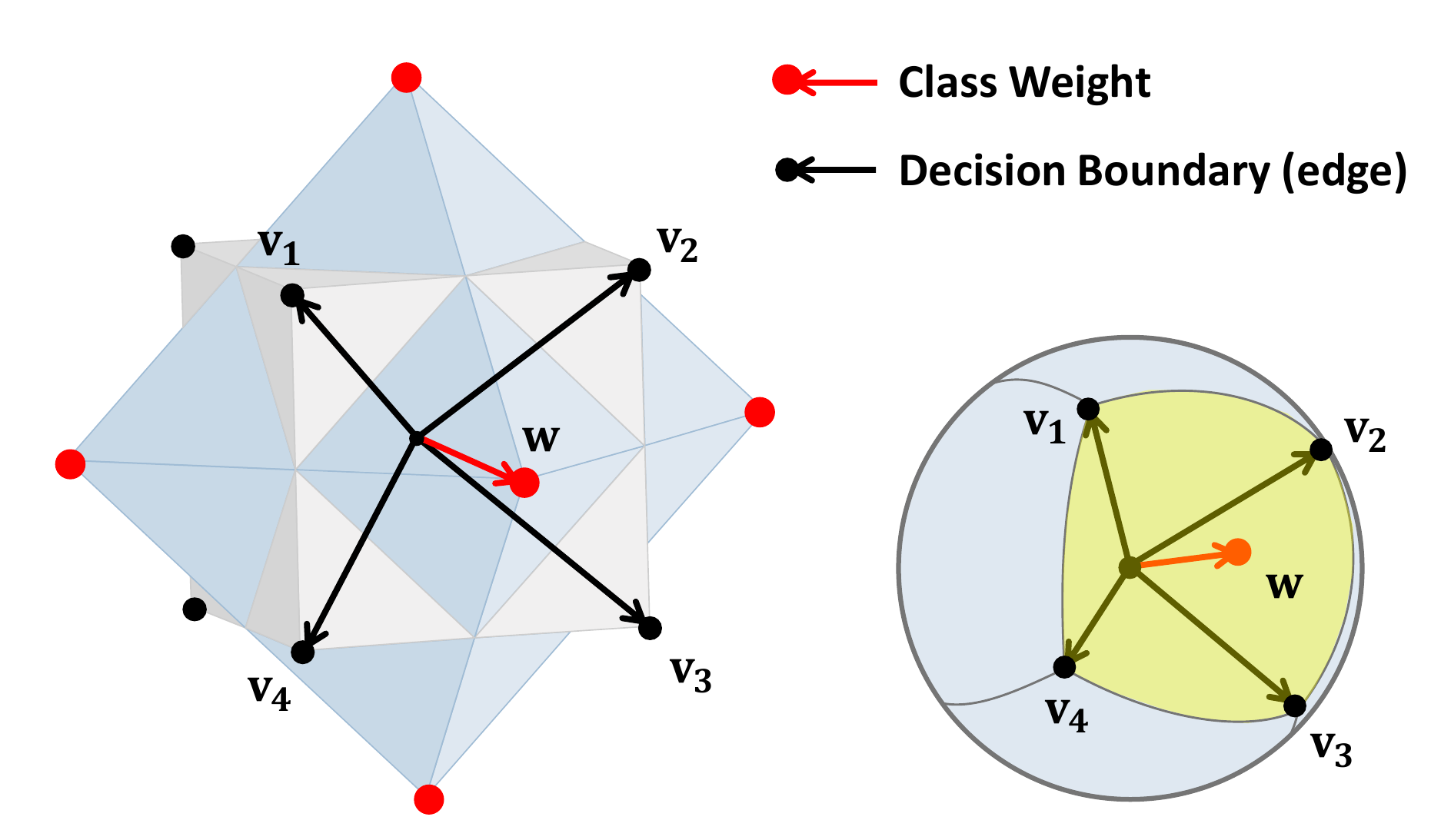}
}
\\
\subfigure[]{
    \includegraphics[width=0.79\columnwidth]{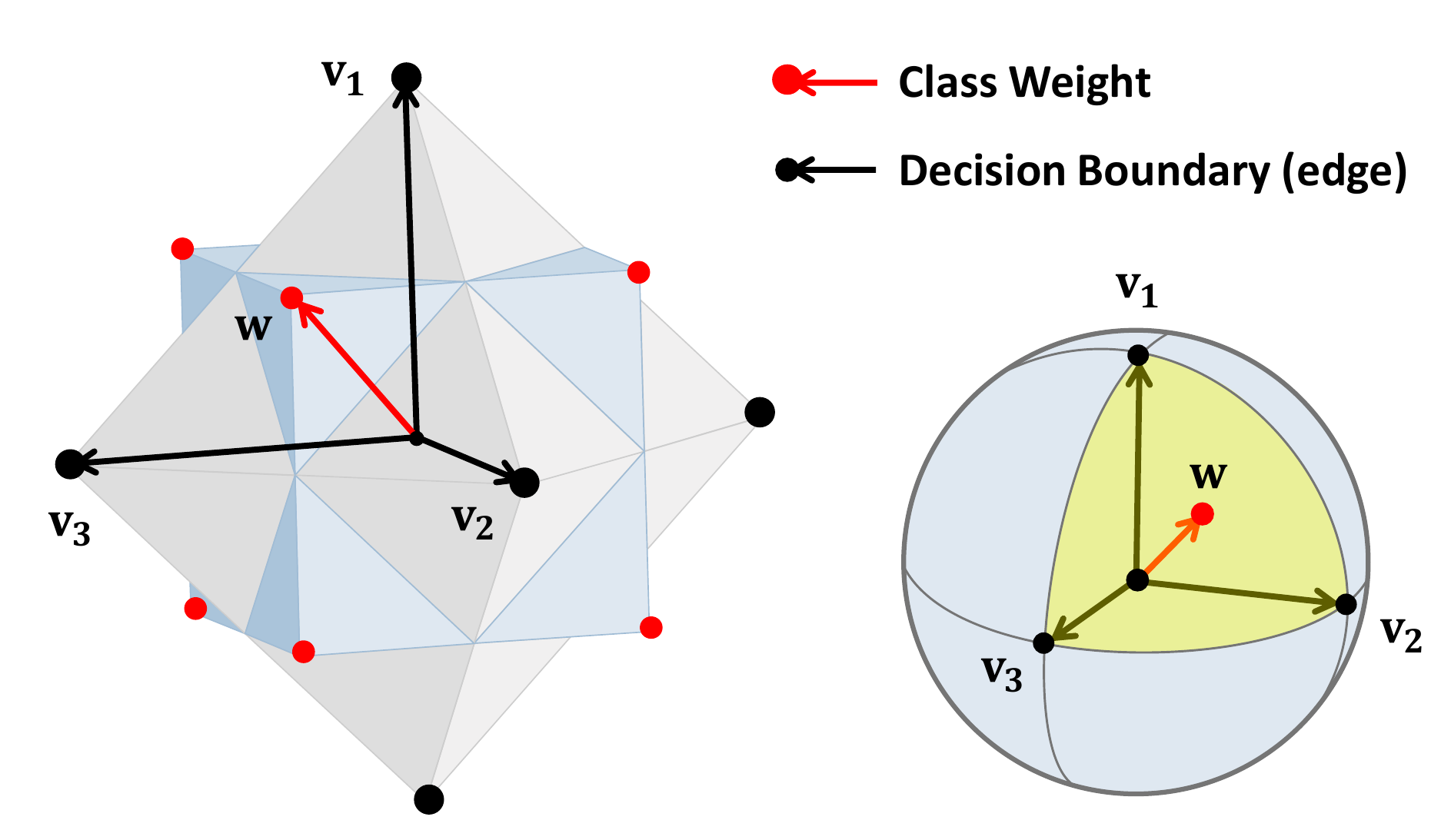}
}
\\
\subfigure[]{
    \includegraphics[width=0.79\columnwidth]{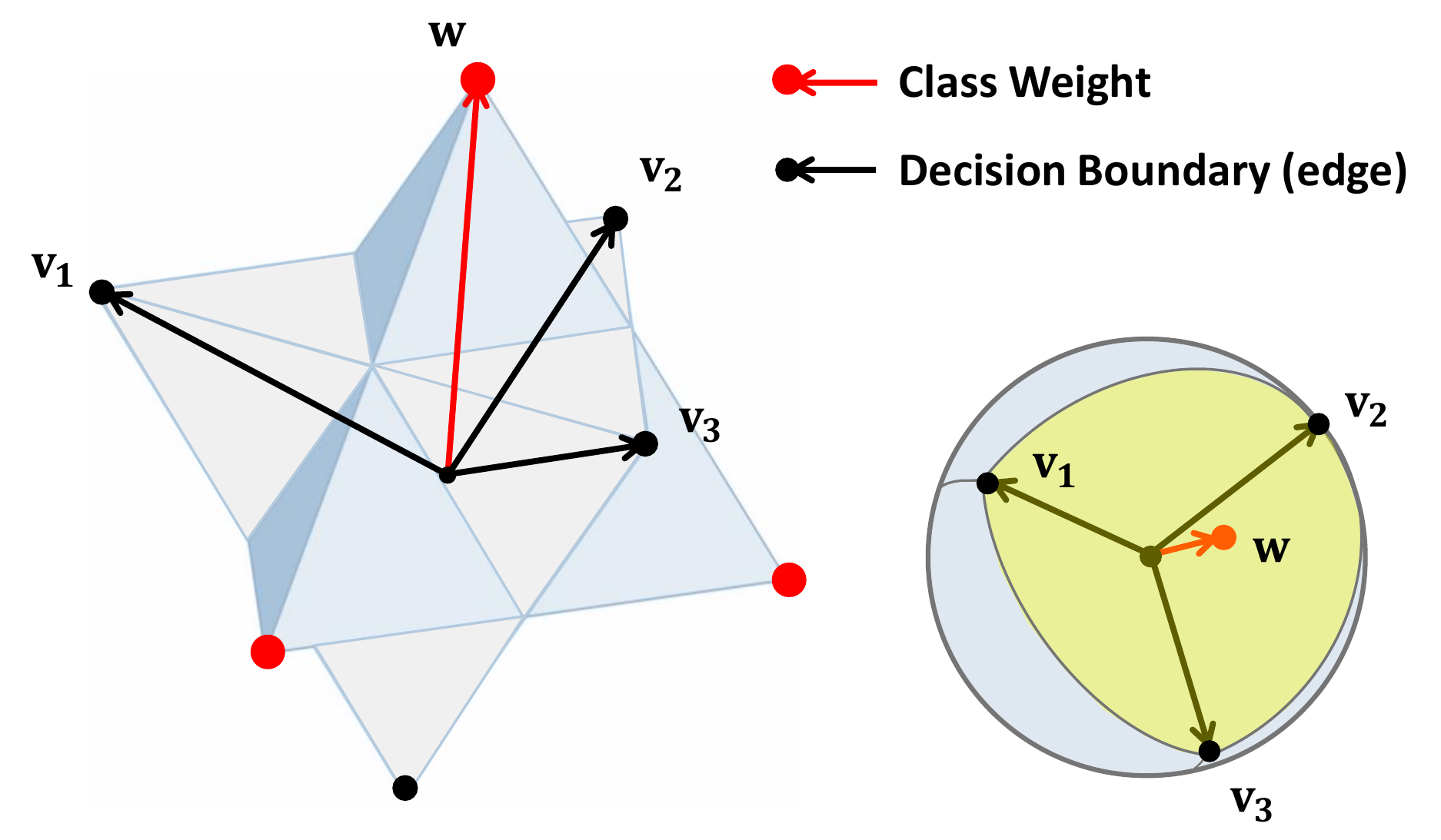}
}
\caption{The RePoNet fixed classifiers decision boundaries in a 3D embedding space: (\emph{a}): $d$-Orthoplex classifier; (\emph{b}):  $d$-Cube classifier; (\emph{c}): $d$-Simplex classifier. On the \emph{left}: the regular polytope classifier (light blue); its dual polytope (grey); a classifier weight $\mathbf{w}$ (red) and its edge decision boundaries $\mathbf{v}_1, \mathbf{v}_2, \dots$~(black). On the \emph{right}: the same entities on the unit sphere. The yellow region indicates where class features are located (only one class weight and the corresponding edge decision boundaries are shown for clarity). The characterization extends to arbitrary dimensions.
}
\label{Duality}
\end{figure}

\subsection{Fixed Classifier Decision Boundaries} 
In binary-classification, the posterior probabilities obtained by softmax in Eq.\ref{softmax_loss_von} are:
\begin{equation}
p_{1}=\frac{\exp ( { \kappa \hat{\mathbf{w}}_{1}^{\top}\hat{\mathbf{f}} })}{\exp ( { \kappa \hat{\mathbf{w}}_{1}^{\top}\hat{\mathbf{f}} })+\exp ( { \kappa \hat{\mathbf{w}}_{2}^{\top}\hat{\mathbf{f}} })}
\end{equation}

\begin{equation}
p_{2}=\frac{\exp ( { \kappa \hat{\mathbf{w}}_{2}^{\top}\hat{\mathbf{f}} })}{\exp ( { \kappa \hat{\mathbf{w}}_{1}^{\top}\hat{\mathbf{f}} })+\exp ( { \kappa \hat{\mathbf{w}}_{2}^{\top}\hat{\mathbf{f}} })}
\end{equation}
where $\mathbf{f}$ is the learned feature vector and $\mathbf{w}_1$ $\mathbf{w}_2$ are the fixed classifier weights. The predicted label will be assigned to the class 1 if $p_{1}>p_{2}$ and to the class 2 if $p_{1}<p_{2}$. By comparing the two probabilities $p_{1}$ and $p_{2},$  $  { \kappa \hat{\mathbf{w}}_{1}^{\top}\hat{\mathbf{f}} } +  { \kappa \hat{\mathbf{w}}_{2}^{\top}\hat{\mathbf{f}} }
$
determines the classification result. The decision boundary is therefore 
$\kappa \hat{\mathbf{w}}_{1}^{\top}\hat{\mathbf{f}} + \kappa \hat{\mathbf{w}}_{2}^{\top}\hat{\mathbf{f}} = 0$. Due to weight normalization the posterior probabilities result in $p_1 = \kappa ||\hat{\mathbf{f}}|| cos(\theta_1)$ and $p_2 = \kappa ||\hat{\mathbf{f}}|| cos(\theta_2)$ and since $p_1$ and $p_2$ share the same feature $\hat{\mathbf{f}}$ the equation $cos(\theta_1)-cos(\theta_2)=0$ is verified at the \emph{angular bisector} between $\mathbf{w}_{1}$ and $\mathbf{w}_{2}$.
Although the above analysis is built on binary-class case, it can be generalized to the multi-class case \cite{Liu2017CVPR}.

In RePoNet angular bisectors define class decision boundaries that follow a symmetry similar to that of the regular polytope defining the classifier. Specifically, the class decision boundaries and the weights of the classifier are related by the duality relationship that holds between regular polytopes. More practically:
\begin{itemize}
\item the set of decision boundaries of the $d$-Simplex classifier is shaped as a $d$-Simplex;
\item the set of the decision boundaries of the $d$-Cube classifier is shaped as a $d$-Orthoplex;
\item the set of the decision boundaries of the $d$-Orthoplex classifier is shaped as a $d$-Cube.
\end{itemize}
Class decision boundaries are still defined as a regular polytope and the features located within such boundaries are therefore maximally separated. The basic intuition behind this result can be better appreciated in 2D exploiting the well known result that all the regular polygons are self-dual \cite{coxeter1963regular}. 
That is, the normal of each side of a regular polygon is parallel to  the direction from the origin towards the vertex of its dual polygon. Fig.~\ref{polygonduality} shows the example introduced in Fig.~\ref{virtual_classes} in which decision boundaries are highlighted with dotted lines according to the dual regular polygon. Fig.~\ref{Duality} illustrates the duality relationship between the weights of the three fixed classifiers proposed and their decision boundaries in the 3D embedding space.

\section{Experimental Results}
We evaluate the correctness (Sec.~\ref{sec_Exchangeability_Assumption}) and the no loss of performance of our approach with respect to standard baselines using trainable and fixed classifiers across a range of datasets and architectures (Sec.~\ref{Generalization and Performance Evaluation}).
All the experiments are conducted with the well known MNIST, FashionMNIST \cite{FashionMNIST2017}, EMNIST \cite{EMNIST17}, CIFAR-10, CIFAR-100 \cite{cifar-10} and ImageNet (ILSVRC2012) \cite{deng2009imagenet} datasets. 
We chose several common CNN architectures (i.e. LeNet, VGG, ResNet, DenseNet), as well as more recent ones (i.e. SeResNeXt50 \cite{Hu_2018_CVPR}, SkResNeXt50 \cite{Li_2019_CVPR} and EfficientNet \cite{DBLP:conf/icml/TanL19})
that have shown to improve performance while maintaining or, in some cases, reducing computational complexity and model size.

\begin{figure}[t]
\hspace{-0.5cm}
    \includegraphics[width=0.5\textwidth]{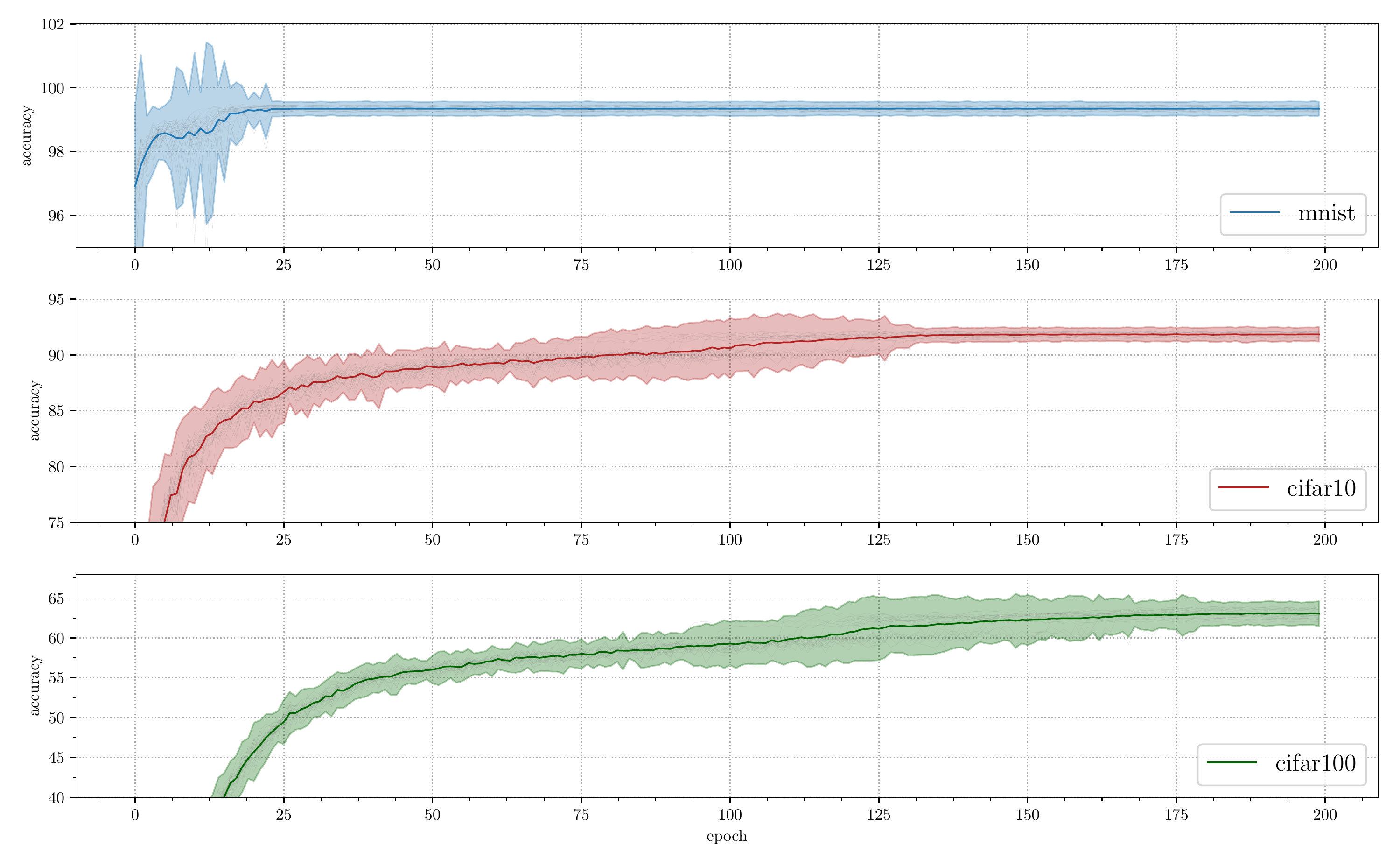}

\caption{Class permutation verification. Average accuracy curves and confidence interval computed from the MNIST, CIFAR-10 and CIFAR-100 datasets (from top to bottom, respectively) under different random permutations of the ground truth labels position.
}
\label{fig_exchangeability}
\end{figure}

\subsection{Hard Permutations Verification}
\label{sec_Exchangeability_Assumption} Since fixed classifiers cannot rely on an adjustable set of subspaces for class feature representation, we verified if some permutations are harder than others for our proposed method. The presence of such hard permutations would preclude the general applicability of our method. 
The standard trainable classifier does not suffer from this problem, when features cannot be well separated trainable classifiers can rearrange the feature subspace directions so that the previous convolutional layers can better disentangle the non-linear interactions between complex data patterns. Instead,  fixed classifiers demand this capability to all the previous layers.

According to this, we generate random permutations of the ground truth label positions\footnote{This is equivalent to randomly permuting the classifier weight vectors set   $\mathbf{W}=\{ \mathbf{w}_j \}_{j=1}^{K}$. } and a new model is learned for each permuted dataset. Fig.~\ref{fig_exchangeability} shows the mean and the $95\%$ confidence interval computed from the accuracy curves of the learned models. To provide further insight into this analysis, 20 out of 500 accuracy curves computed for each dataset are also shown. Specifically, the evaluation is performed on three different datasets with an increasing level of complexity (i.e MNIST, CIFAR-10 and CIFAR-100).  
All the models are trained for 200 epochs to make sure that the models trained with CIFAR-100 achieve convergence. 

\begin{figure}[t]
\vspace{-0.2cm}
\centering
\hspace{-0.65cm}
\subfigure[]{\label{fig_even_odd:a}
\includegraphics[height=0.47\linewidth,valign=t]{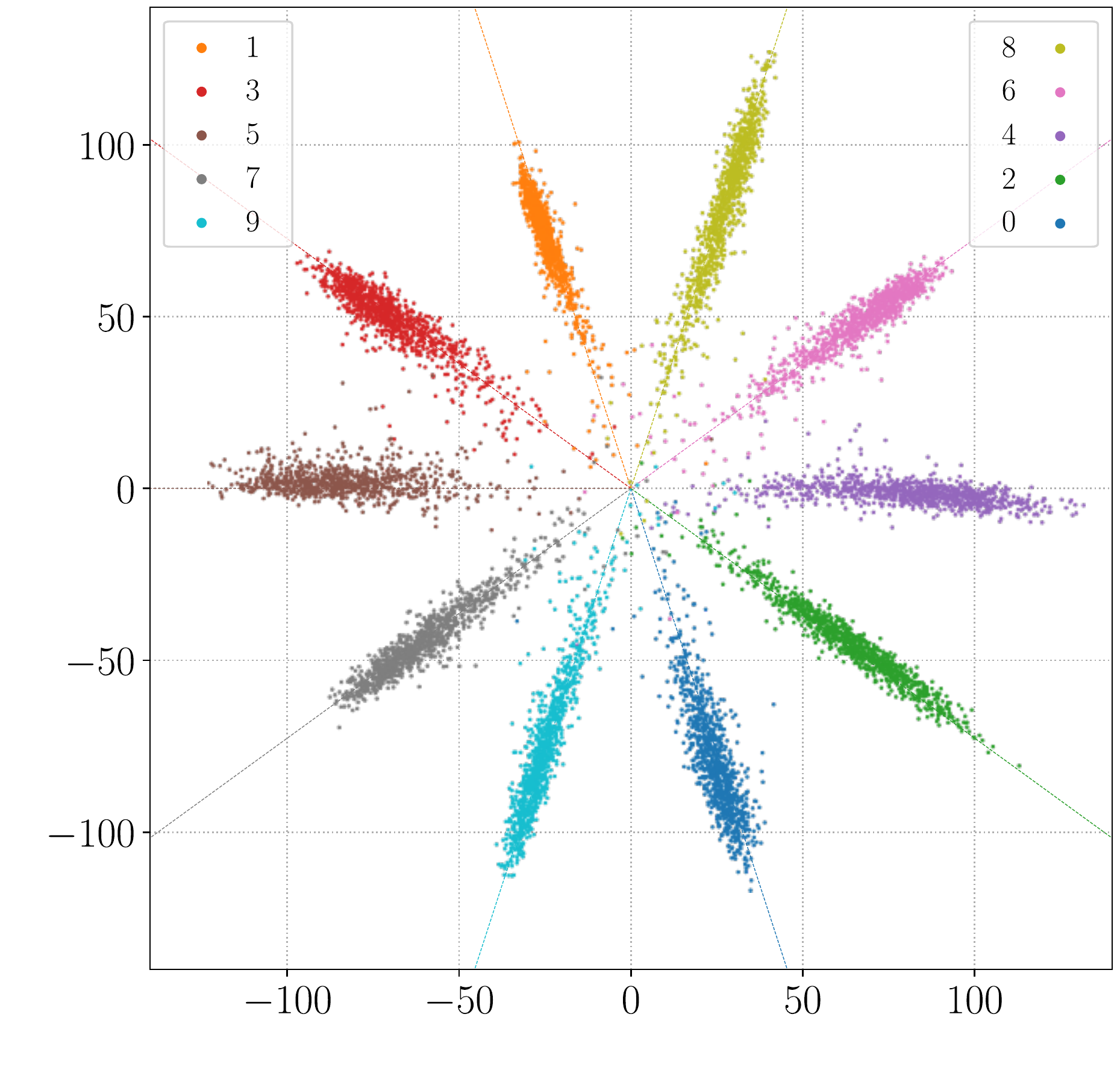}
}
\subfigure[]{\label{fig_CIFAR10:b}
\includegraphics[height=0.47\linewidth,valign=t]{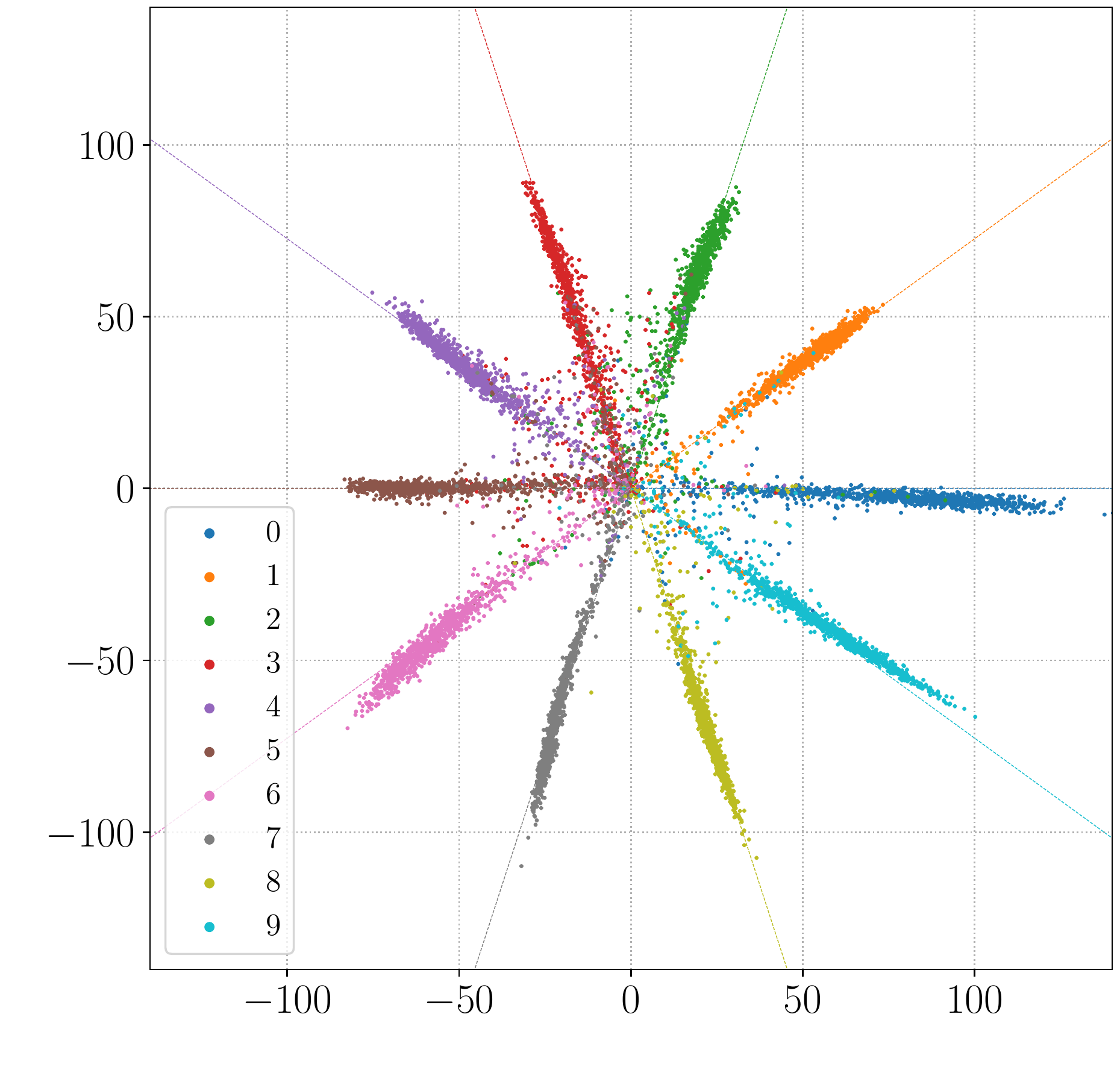} 
}
\caption{ The distribution of features learned using a 10-sided regular polygon. \emph{(a)}: A special permutation of classes is shown in which the MNIST even and odd digits are placed in the positive and negative half-space of the abscissa respectively. \emph{(b)}: The features learned using the CIFAR-10 dataset.
}
\label{fig_even_odd}
\end{figure}

In order to address the most severe possible outcomes that may happen, for this experiment we used the $d$-Cube fixed classifier. Being the hardest to optimize, this experiment can be regarded as a worst case analysis scenario for our method. As shown in the same figure, the performance is substantially insensitive to both permutations and datasets. The average reduction in performance at the end of the training process is negligible and the confidence intervals reflect the complexity of the datasets. Although the space of permutations cannot be exhaustively evaluated even for a small number of classes, we have achieved proper convergence for the whole set of 1500 learned models. 
The experiment took 5 days on a \mbox{Nvidia DGX-1}.

On the basis of this evidence, we can conclude that fixing the classifier (therefore not having access to a set of adjustable subspaces for class feature representation) does not affect the expressive power of neural networks.
This experiment also provides a novel and more systematic empirical evidence of the general applicability and correctness of fixed classifiers
with respect to \cite{hoffer2018fix} where only one permutation was tested. 

\begin{table*}
\caption{Reported accuracy (\%) of the RePoNet method on \textsc{MNIST}, \textsc{EMNIST}, \textsc{FashionMNIST} datasets on different combinations of architectures and relative learned classifier baselines.
}
	\centering
	\footnotesize 
   \scalebox{1.00} {

\begin{tabular}{cccc}
  \toprule

  \multicolumn{1}{c}{} &
  \thead{\textsc{MNIST} \\[-2px] \scriptsize ($K=10$)}  & \thead{\textsc{EMNIST} \\[-2px] \scriptsize ($K=47$) } & \thead{ \textsc{FashionMNIST} \\[-2px] \scriptsize ($K=10$) }\\

  \cmidrule(lr){2-4}
  \multicolumn{1}{c}{\thead{\textsc{Architecture}}}  &
  \multicolumn{3}{c}{ \thead{\textsc{LeNet++}}}   \\
  
  \midrule
  RePoNet $K$-sided-polygon                           & $99.24_{d=2}$     & $72.81_{d=2}$     & $92.48_{d=2}$          \\
  Hadamard fixed classifier \cite{hoffer2018fix}       & $21.14_{d=2}$     & $4.12_{d=2}$      & $19.89_{d=2}$          \\ 
  Learned Classifier                         		   & $99.21_{d=2}$     & $73.08_{d=2}$     & $92.79_{d=2}$          \\

  \midrule
  RePoNet $d$-Cube                      			  & $99.58_{d=4}$     & $88.12_{d=6}$     & $94.01_{d=4}$           \\
  Hadamard fixed classifier \cite{hoffer2018fix}      & $41.99_{d=4}$     & $15.12_{d=6}$     & $37.16_{d=4}$           \\
    Learned Classifier                         		  & $99.41_{d=4}$     & $86.96_{d=6}$     & $93.94_{d=4}$           \\

  \midrule
  RePoNet $d$-Orthoplex                              & $99.66_{d=5}$     & $88.19_{d=24}$    & $94.84_{d=5}$           \\
  Hadamard fixed classifier \cite{hoffer2018fix}      & $79.34_{d=5}$     & $60.34_{d=24}$    & $74.22_{d=5}$           \\
  Learned Classifier                         		  & $99.07_{d=5}$     & $87.66_{d=24}$    & $94.21_{d=5}$           \\

  \midrule
  RePoNet $d$-Simplex                                & $99.71_{d=9}$     & $88.89_{d=46}$    & $94.29_{d=9}$           \\
  Hadamard fixed classifier \cite{hoffer2018fix}      & $99.12_{d=9}$     & $88.48_{d=46}$    & $94.30_{d=9}$           \\ 
  Learned classifier                         		  & $99.41_{d=9}$     & $88.33_{d=46}$    & $94.41_{d=9}$           \\

  \midrule
  Hadamard fixed classifier \cite{hoffer2018fix}      & $99.54_{d=512}$   & $88.35_{d=512}$   & $94.14_{d=512}$ \\ 
  Learned classifier                         		  & $99.29_{d=512}$   & $88.87_{d=512}$   & $94.28_{d=512}$ \\

  \bottomrule
\end{tabular}
} 
\label{table:lenetpp}
\end{table*}
\setlength{\extrarowheight}{2pt} 
\begin{table*}
\caption{
Reported accuracy (\%) of the RePoNet method on the \textsc{CIFAR-10} dataset on different combinations of architectures and relative baselines. }
	\centering
	\footnotesize 
   \scalebox{1.00} {

\sisetup{detect-weight=true,detect-inline-weight=math,table-number-alignment=center,round-precision=2,round-integer-to-decimal=false, round-mode=places}
\begin{tabular}{
    cccccc
  }
  \toprule
  \multicolumn{1}{c}{} &
  \multicolumn{5}{c}{ \thead{CIFAR-10 \scriptsize ($K=10$)}  }
  \\

  \cmidrule(lr){2-6} \multicolumn{1}{c}{\thead{\textsc{Architecture}}}  &
  \textsc{VGG13} & \textsc{VGG19} & \textsc{ResNet50} & \textsc{SENet18} & \textsc{DenseNet169} \\
  \scriptsize{Training hyperparameters} & \scriptsize{\textsc{adam}} & \scriptsize{\textsc{adam}}  & \scriptsize{\textsc{sgd}}  & \scriptsize{\textsc{sgd}} & \scriptsize{\textsc{sgd}} \\
  \midrule
  RePoNet $K$-sided-polygon                          & $90.79_{d=2}$     & $91.54_{d=2}$     & $92.78_{d=2}$         & $92.63_{d=2}$     & $92.74_{d=2}$        \\
  Hadamard fixed classifier \cite{hoffer2018fix}      & $19.45_{d=2}$     & $19.19_{d=2}$     & $19.69_{d=2}$         & $19.77_{d=2}$     & $19.76_{d=2}$        \\
  Learned classifier                         		  & $90.41_{d=2}$     & $91.17_{d=2}$     & $93.15_{d=2}$         & $93.25_{d=2}$     & $92.89_{d=2}$        \\
  \midrule
  RePoNet $d$-Cube                      			  & $92.26_{d=4}$     & $92.58_{d=4}$     & $94.86_{d=4}$         & $94.96_{d=4}$     & $93.94_{d=4}$        \\
  Hadamard fixed classifier \cite{hoffer2018fix}      & $37.19_{d=4}$     & $36.95_{d=4}$     & $37.89_{d=4}$         & $38.05_{d=4}$     & $38.12_{d=4}$        \\
    Learned classifier                         		  & $92.14_{d=4}$     & $92.21_{d=4}$     & $95.03_{d=4}$         & $94.95_{d=4}$     & $94.97_{d=4}$        \\
  \midrule
  RePoNet $d$-Orthoplex                              & $92.51_{d=5}$     & $92.47_{d=5}$     & $95.25_{d=5}$         & $95.05_{d=5}$     & $95.16_{d=5}$        \\
  Hadamard fixed classifier \cite{hoffer2018fix}      & $73.77_{d=5}$     & $72.46_{d=5}$     & $75.99_{d=5}$         & $75.95_{d=5}$     & $75.73_{d=5}$        \\
  Learned classifier                         		  & $92.28_{d=5}$     & $92.21_{d=5}$     & $95.18_{d=5}$         & $95.08_{d=5}$     & $95.41_{d=5}$        \\
  \midrule
  RePoNet $d$-Simplex                                & $92.71_{d=9}$     & $92.59_{d=9}$     & $95.66_{d=9}$         & $95.36_{d=9}$     & $95.32_{d=9}$        \\
  Hadamard fixed classifier \cite{hoffer2018fix}      & $92.03_{d=9}$     & $92.37_{d=9}$     & $95.53_{d=9}$         & $95.25_{d=9}$     & $94.92_{d=9}$        \\
  Learned classifier                         		  & $91.89_{d=9}$     & $92.60_{d=9}$     & $95.08_{d=9}$         & $95.20_{d=9}$     & $95.32_{d=9}$        \\
  \midrule
  Hadamard fixed classifier \cite{hoffer2018fix}      & $90.11_{d=512}$   & $88.32_{d=512}$   & $95.36_{d=512}$       & $95.49_{d=512}$   & $95.68_{d=512}$      \\
  Learned classifier                         		  & $92.34_{d=512}$   & $92.42_{d=512}$   & $95.53_{d=512}$       & $95.26_{d=512}$   & $95.68_{d=512}$      \\

  \bottomrule
\end{tabular}

} 
\label{table:cifar10}
\end{table*}

We finally report qualitative results of a learned permuted dataset. Fig.~\ref{fig_even_odd}(a) shows features learned in a $k$-sided polygon (2d embedding space) on the MNIST dataset. In particular the model is learned with a special permutation of the labels (manually selected) that places even and odd digits features respectively on the positive and negative half space of the abscissa. 
Fig.~\ref{fig_even_odd}(b) shows the features of on \textsc{CIFAR-10} learned with a similar $10$-sided-polygon. It can be noticed that features are distributed following the same polygonal pattern shown in Fig.~\ref{fig_even_odd}(a).

\begin{table*}
\caption{
Reported accuracy (\%) of the RePoNet method on the \textsc{CIFAR-100} dataset on different combinations of architectures and relative baselines.
}
	\centering
	\footnotesize 
   \scalebox{1.00} {

\begin{tabular}{cccccc}
  \toprule

  \multicolumn{1}{c}{} &
  \multicolumn{5}{c}{ \thead{CIFAR-100 \scriptsize ($K=100$)}}
  \\

  \cmidrule(lr){2-6} \multicolumn{1}{c}{\thead{\textsc{Architecture}}}  &
  \textsc{VGG13} & \textsc{VGG19} & \textsc{ResNet50} & \textsc{SENet18} & \textsc{DenseNet169} \\
  
  \scriptsize{Training hyperparameters} & \scriptsize{\textsc{adam}} & \scriptsize{\textsc{adam}}  & \scriptsize{\textsc{sgd}}  & \scriptsize{\textsc{sgd}} & \scriptsize{\textsc{sgd}} \\
  \midrule
  RePoNet $K$-sided-polygon                           & $36.22_{d=2}$     & $37.65_{d=2}$     & $33.39_{d=2}$        & $35.26_{d=2}$     & $30.04_{d=2}$           \\
  Hadamard fixed classifier \cite{hoffer2018fix}       & $ 1.75_{d=2}$     & $ 1.75_{d=2}$     & $1.61_{d=2}$        & $1.80_{d=2}$     & $1.64_{d=2}$           \\
  Learned classifier                         		   & $37.56_{d=2}$     & $35.83_{d=2}$     & $33.30_{d=2}$        & $40.57_{d=2}$     & $32.87_{d=2}$           \\

  \midrule
  RePoNet $d$-Cube                      			  & $64.35_{d=7}$     & $65.32_{d=7}$     & $67.27_{d=7}$         & $69.38_{d=7}$     & $68.99_{d=7}$           \\
  Hadamard fixed classifier \cite{hoffer2018fix}      & $ 5.96_{d=7}$     & $ 5.52_{d=7}$     & $5.91_{d=7}$         & $6.27_{d=7}$     & $6.08_{d=7}$           \\
    Learned classifier                         		  & $64.11_{d=7}$     & $65.29_{d=7}$     & $74.96_{d=7}$         & $75.29_{d=7}$     & $75.51_{d=7}$           \\

  \midrule
  RePoNet $d$-Orthoplex                              & $68.78_{d=50}$    & $69.76_{d=50}$    & $78.23_{d=50}$        & $77.24_{d=50}$    & $79.41_{d=50}$           \\
  Hadamard fixed classifier \cite{hoffer2018fix}      & $43.88_{d=50}$    & $43.89_{d=50}$    & $50.33_{d=50}$        & $49.56_{d=50}$    & $50.65_{d=50}$           \\
  Learned classifier                         		  & $68.13_{d=50}$    & $68.41_{d=50}$    & $78.22_{d=50}$        & $77.15_{d=50}$    & $78.83_{d=50}$           \\

  \midrule
  RePoNet $d$-Simplex                                & $68.61_{d=99}$    & $68.69_{d=99}$    & $79.02_{d=99}$        & $78.20_{d=99}$    & $80.01_{d=99}$           \\
  Hadamard fixed classifier \cite{hoffer2018fix}      & $67.23_{d=99}$    & $67.18_{d=99}$    & $78.82_{d=99}$        & $77.21_{d=99}$    & $79.41_{d=99}$           \\
  Learned classifier                         		  & $68.15_{d=99}$    & $68.87_{d=99}$    & $78.58_{d=99}$        & $77.42_{d=99}$    & $79.05_{d=99}$           \\

  \midrule
  Hadamard fixed classifier \cite{hoffer2018fix}      & $63.16_{d=512}$   & $64.46_{d=512}$    & $78.78_{d=512}$      & $77.94_{d=512}$   & $79.44_{d=512}$           \\
  Learned classifier                         		  & $68.56_{d=512}$   & $68.47_{d=512}$    & $77.96_{d=512}$      & $77.63_{d=512}$   & $79.63_{d=512}$           \\

  \bottomrule
\end{tabular}
} 
\label{table:cifar100}
\end{table*}

\subsection{Generalization and Performance Evaluation}
\label{Generalization and Performance Evaluation}
Having verified that the order position of the class labels does not adversely affect the proposed method, in this section we evaluate the classification performance of RePoNet on the following datasets: MNIST, EMNIST, FashionMNIST, CIFAR-10, CIFAR-100 and ImageNet. 
The RePoNet method is compared with CNN baselines with learned classifiers and the fixed classifier method reported in \cite{hoffer2018fix}, that has been implemented for different architectures and different dimensions of the embedding space. Except for the final fixed classifier all the compared methods have exactly the same architecture and training settings as the one that RePoNet uses. 
\\

\subsubsection{\textbf{{MNIST and CIFAR}}} 
We trained the so called LeNet++ architecture \cite{wen2016discriminative} on all the MNIST family datasets. The network is a modification of the LeNet \cite{lecun1998gradient} to a deeper and wider network including parametric rectifier linear units (pReLU) \cite{he2015delving}.
For the evaluation on the CIFAR-10 and CIFAR-100 datasets, we further trained VGG \cite{SimonyanZ14a} with depth 13 and 19, ResNet50 \cite{he2016deep}, 
SeNet \cite{hu2018squeeze} and DenseNet169 \cite{Densenet2017}. 
Popular network architectures for ImageNet require modifications to adapt to the CIFAR 32x32 input size. According to this, our experiments follow the publicly available implementations\footnote{
\url{https://github.com/bearpaw/pytorch-classification} and \url{https://github.com/kuangliu/pytorch-cifar} }. 
We compared all the variants of our approach for each architecture including trainable classifiers with different dimensions of the feature space. 
The mini batch size is 256 for both the MNIST family datasets and the CIFAR-10/100 datasets.
For the CIFAR datasets, we compared both a ``vanilla'' learning setup with no hyperparameters tuning based on the Adam optimizer (learning rate $0.0005$) for the VGG architectures and a learning setup based on SGD with a specific learning rate schedule (starting from 0.1 and decreasing by a factor of 10 after 150 and 250 epochs) for ResNet50, SEnet18 and DenseNet169 architectures.
As hyperparameters tuning is an integral part of Deep Learning we provided two opposite learning setup.
\\
\indent
Test-set accuracy for this experiment is reported in Tab.~\ref{table:lenetpp}, \ref{table:cifar10} and \ref{table:cifar100} for MNISTs, CIFAR-10 and CIFAR-100, respectively.
In addition to the well-known MNIST and FashionMnist, we included EMNIST dataset having $47$ classes including lower/upper case letters and digits. This allows to quantify with a specific dataset and architecture, as in CIFAR-10 and CIFAR-100, the classification accuracy with a higher number of classes. 
Each entry in the tables report the test-set accuracy. The subscript indicates the specific feature space dimension $d$ used. The results reveal and confirm that the RePoNet method achieves comparable classification accuracy of other trainable classifier models. This evidence is in agreement on all the combinations of datasets, architectures, number of classes and feature space dimensions considered. All the RePoNet variants exhibit similar behavior even in hard combinations such as the \textsc{CIFAR-100} dataset in a low dimensional feature space. For example, the RePoNet $d$-Cube fixed classifier implemented with the VGG19 architecture achieves an accuracy of $65.32\%$ in a $d=7$ dimensional feature space. A fully trainable classifier in a feature space of dimension $d=512$ (i.e. two orders of magnitude larger), achieves a moderate improvement of about $3\%$ ($68.47\%$). On the other hand, with a significantly lower feature dimension of $d=50$, RePoNet $d$-Orthoplex improves the accuracy to $69.76\%$.
All the RePoNet variants exhibit similar behavior also in the case of more sophisticated architectures trained with SGD scheduled learning rates to match state-of-the-art performance. 
RePoNet classifiers are both agnostic to architectures and training setup and are able to improve accuracy similar to trainable classifiers.

Results also show that the Hadamard fixed classifier \cite{hoffer2018fix} does not succeed to learn when the number of classes is larger than the number of unique weight directions in the embedding space (i.e. $d < K$). As expected, this effect is present for simple datasets as the MNIST digits dataset, however as reported in \cite{hoffer2018fix} Section 4.2 (Possible Caveats) as the number of classes $K$ increases the effect is less pronounced. 

When $d \! \approx \! K$ or $d \! > \! K$, classification performance is similar. However, as shown in Fig.~\ref{fig_learning_speed}(a) 
RePoNet converges faster than \cite{hoffer2018fix}, and with the same speed as the trainable baselines. Our conjecture is that with our symmetrical fixed classifiers, each term in the loss function tends to have the same magnitude centered around the mean of the distribution (i.e. the von Mises-Fisher distribution is similar to the Normal distribution) and therefore the average computed in the loss is a good estimator. Instead, in the Hadamard classifier the terms may have different magnitudes and  ``important'' errors in the loss may not be taken into account correctly by simple averaging.

\subsubsection{\textbf{ImageNet}}
Finally, we evaluated our method
on the 1000 object category classification problem defined by the ImageNet dataset. This dataset consists of a 1.2M image training set and a 100k image test set.
We compared all the variants of our approach on different combinations of architectures and their relative trainable classifiers. The comparison also includes the Hadamard classifier. 
\\
\indent
Experiments have been conducted in two different configurations of the training hyperparameters.
\emph{First}, we performed experiments using the Adam optimizer and simple augmentation based on random cropping and horizontal flipping on well-established networks such as ResNet50 \cite{he2016deep} and DenseNet169 \cite{Densenet2017}. The learning rate is automatically adjusted when a plateau in model performance is detected.
We trained for $250$ epochs with batch size $64$ with an initial learning rate of $0.0005$.
With this configuration, we aim to evaluate our method without performing any specific hyperparameter optimization or exploiting large computational resources.
\emph{Second}, we evaluated our method with more sophisticated CNN architectures, namely SKresNeXt, SEresNeXt, and EfficientNet with related training hyperparameters. With this configuration, the aim is to evaluate whether our method can reach state-of-the-art performance.
The SKresNeXt and SEresNeXt architectures integrate the SE and SK blocks, \cite{hu2018squeeze} and \cite{li2019selective} respectively, with the ResNeXt architecture  \cite{xie2017aggregated}. The benefit of these variants is to maintain computational complexity and model size similar to the SEnet and SKnet architectures while further improving performance. The third architecture, EfficientNet \cite{DBLP:conf/icml/TanL19}, achieves state-of-the-art performance using significantly fewer parameters than other state-of-the-art models. 
As these architectures typically require a
large effort to tune the training hyperparameters, we trained our method on top of these models following the settings reported in the original papers. 
Specifically, we train EfficentNet-B2 following \cite{DBLP:conf/icml/TanL19}: RMSProp optimizer with decay 0.9 and momentum 0.9; batch norm momentum 0.99; initial learning rate 0.256 that decays by 0.97 every 2.4 epochs; weight decay 1e-5.
Analogously, SKresNeXt50 and SEresNeXt50 are trained following the ResNeXt50 \cite{xie2017aggregated}: SGD optimizer, weight decay 0.0001; momentum 0.9; initial learning rate of 0.1, divided by 10 for three times using a specific schedule reported in the paper.
For all the three models we used automated data augmentation techniques from \cite{cubuk2020randaugment} (RandAugment) with distortion magnitude $7$. SeResNeXt50 and SkResNeXt50 were trained for $250$ epochs with $192$ batch size. EfficientNet-B2 was trained for $450$ epochs with $120$ batch size. Our evaluation is based on the pytorch-image-models\footnote{\url{https://github.com/rwightman/pytorch-image-models}} repository. 

\begin{table*}
\caption{
Reported accuracy (\%) of the RePoNet method on the \textsc{ImageNet} dataset on different combinations of architectures and relative classifier baselines.}
\centering
\begin{tabular}{ccccccc} \hline

  \toprule

  \textsc{Architecture}             & \textsc{ResNet50} & \textsc{DenseNet169}  & \textsc{SeResNeXt50}  & \textsc{SkResNeXt50} & \textsc{EfficientNet-B2} \\
  \scriptsize{Training hyperparameters}  & \scriptsize{\textsc{adam}} & \scriptsize{\textsc{adam}}  & \scriptsize{\textsc{sgd}+\textsc{r}and\textsc{a}ug}  & \scriptsize{\textsc{sgd}+\textsc{r}and\textsc{a}ug} & \scriptsize{\textsc{rmsprop}+\textsc{r}and\textsc{a}ug} \\
  \midrule

  RePoNet $d$-Cube                 & $63.44_{d=10}$    & $63.63_{d=10}$    & $73.58_{d=10}$    & $74.80_{d=10}$ & $75.62_{d=10}$ \\
  Learned classifier      & $68.82_{d=10}$    & $68.03_{d=10}$    & $76.66_{d=10}$    & $77.49_{d=10}$ & $77.42_{d=10}$ \\

  \midrule
  RePoNet $d$-Orthoplex            & $73.71_{d=500}$   & $74.20_{d=500}$   & $79.95_{d=500}$    & $79.66_{d=500}$ & $80.07_{d=500}$ \\
  Learned classifier      & $73.67_{d=500}$   & $73.70_{d=500}$   & $77.60_{d=500}$    & $80.18_{d=500}$ & $79.27_{d=500}$ \\

  \midrule
  RePoNet $d$-Simplex              & $74.13_{d=999}$   & $74.03_{d=999}$   & $80.25_{d=999}$    & $80.17_{d=999}$ & $80.61_{d=999}$ \\
  Learned classifier      & $73.96_{d=999}$   & $73.37_{d=999}$   & $77.99_{d=999}$    & $80.08_{d=999}$ & $79.36_{d=999}$ \\

  \midrule

  Hadamard fixed classifier \cite{hoffer2018fix} & $74.07_{d=2048}$   & $73.95_{d=1669}$ & $80.25_{d=2048}$ & $80.19_{d=2048}$ & $79.74_{d=1408}$\\

  Learned classifier & $74.11_{d=2048}$   & $74.01_{d=1669}$ & $79.95_{d=2048}$ & $80.09_{d=2048}$ & $80.57_{d=1408}$\\
  \bottomrule

\end{tabular}
\vspace{0.08cm}
\\
\label{table:imagenet}
\end{table*}

Tab.~\ref{table:imagenet} summarizes our results.
As can be clearly noticed, except for the $d$-Cube there is no substantial difference between the performance of our fixed classifiers and the learned classifiers. This holds also in the case of the learned classifiers in their original architecture implementation (shown in the bottom line of the Table). The table also shows that RePoNet accuracy is comparable with the Hadamard fixed classifier \cite{hoffer2018fix}. 
As in the cases of CIFAR-10 and CIFAR-100, also with ImageNet the  accuracy of the \mbox{$d$-Cube} is lower than the corresponding learned classifiers. We argue this is mainly due to the difficulty of performing optimization in the $d=10$ dimensional space 
due to the fact that the angle between each class weight vector and its $d$ adjacent weight vectors approaches to zero as the dimension increases (Fig.~\ref{fig:anglefcn}). 
However, 
the $d$-Cube classifier shows the largest relative improvement as the representational power of the  architecture increases (left to right). 
For example, the accuracy of the  \mbox{$d$-Cube-\textsc{EfficientNet-B2}} fixed classifier is $12.18$ percentage points larger than the \mbox{$d$-Cube-\textsc{ResNet50}} (i.e. $75.62-63.44 = 12.18$). This relative performance improvement is substantially higher than that of the corresponding learned classifier (i.e. $77.42 - 68.82 = 8.6$). This result is quantitatively consistent with the underlying assumption of this paper and provides further support on the fact that the adjustable capability of the final classifier can be successfully demanded to previous layers. 
The other two RePoNet variants substantially achieve the same accuracy of the learned classifiers, irrespective whether they have similar ($d= \{ 10, 500, 999 \}$) or higher feature space dimension ($d= \{ 2048, 1669, 1408 \}$) as in their original architecture implementations. They do not show sensible relative performance improvement with increasing representational power of the network. More importantly, both the $d$-Simplex and the $d$-Orthoplex classifiers reach state-of-the-art accuracy (around 80\%) when combined with competitive architectures. This confirms the validity and the absence of a loss of generalization  of our method.

\begin{table}[b]
    \caption{The number of parameters of each network and the percentage (\%) of saved parameters on the ImageNet dataset ($d$ indicates the feature dimension).}
    \centering
    \resizebox{.49\textwidth}{!}{%
    \begin{tabular}{lcccc}
    \toprule
    & &   & \textsc{Saved Params ( \% )} &   \\
    \cmidrule(lr){3-5}
    \textsc{Architecture} & \textsc{Param\#} & \textsc{$d$-Cube}  & \textsc{$d$-Orthoplex} & \textsc{$d$-Simplex} \\
    \midrule
    DenseNet169 & 14.15M & $11.65_{d=10}$ & $5.89_{d=500}$      & $0.02_{d=999}$ \\
    ResNet50 & 25.56M & $7.94_{d=10}$  & $4.01_{d=500}$      & $0.01_{d=999}$ \\
    SeResNeXt50 & 27.56M & $7.36_{d=10}$  & $3.72_{d=500}$      & $0.01_{d=999}$ \\
    SkResNeXt50 & 27.50M & $7.38_{d=10}$  & $3.73_{d=500}$      & $0.01_{d=999}$ \\
    EfficientNet-B2 & 9.11M & $15.31_{d=10}$  & $7.74_{d=500}$      & $0.03_{d=999}$ \\
    \bottomrule
    \end{tabular}%
    }
    \label{params1}
    \label{params2}
\end{table}

Finally, Tab.~\ref{params2} shows the total number of parameters for each network in comparison with their original learned classifiers (i.e. bottom line in Tab.~\ref{table:imagenet}). The $d$-Orthoplex-\textsc{EfficinetNet-B2} fixed classifier saves $7.74\%$ of the network parameters while achieving the same accuracy (around 80\%). 
It is worth to notice that the $d$-Cube-\textsc{EfficinetNet-B2} with 7.7M of parameters ($15.31\%$ savings) achieves similar accuracy of a vanilla ResNet50 baseline (i.e. around 75\% accuracy) having 25.5M of parameters. 
\subsection{Training Time}
\begin{figure}[t]
    \centering
    \subfigure[]{
        \hspace{-0.5cm}
        \includegraphics[width=0.99\columnwidth]{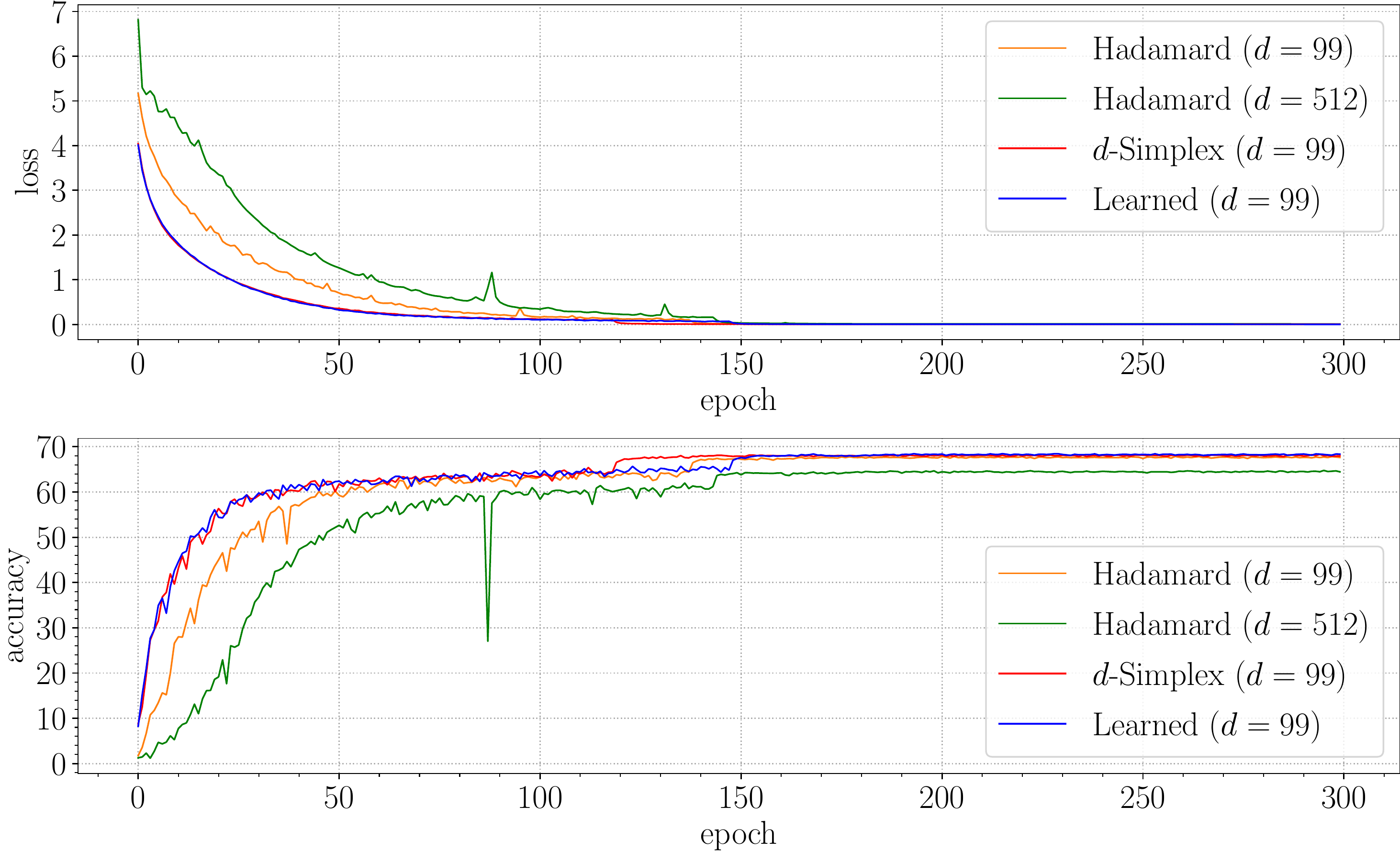}
        \vspace{.2cm}
    }
    \subfigure[]{
        \hspace{-0.5cm}
        \vspace{0.1cm}
        \includegraphics[width=0.49\textwidth]{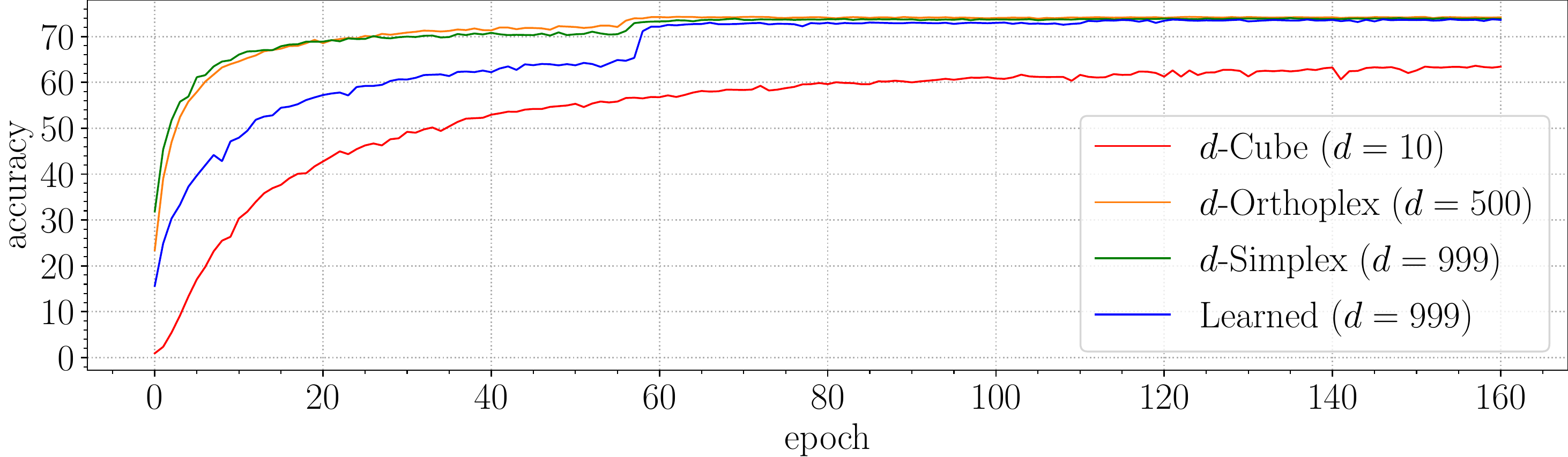}
    }
        \vspace{-0.09cm}
    \caption{Speed of convergence comparison (\textsc{Adam}). \emph{(a)}: Training error curves (\emph{top}) and test accuracy curves (\emph{bottom}) using the CIFAR-100 dataset with the VGG19 architecture. \emph{(b)}: ImageNet learning speed using DenseNet169. As evidenced from the figures, the proposed method has faster convergence.
    }
    \label{fig_learning_speed}
\end{figure}

\begin{figure}[t]
    \centering
    \vspace{.02cm}
    \subfigure[]{
        \hspace{-0.5cm}
        \includegraphics[width=0.49\textwidth]{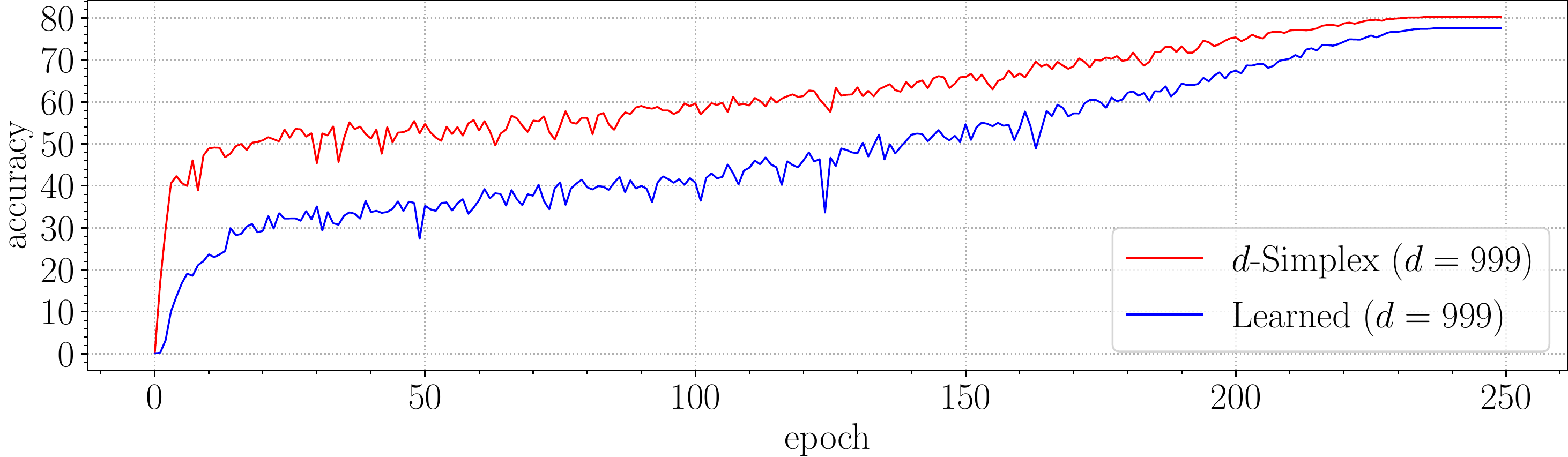}
    \vspace{-.1cm}
    }
    \subfigure[]{
        \hspace{-0.5cm}
        \includegraphics[width=0.49\textwidth]{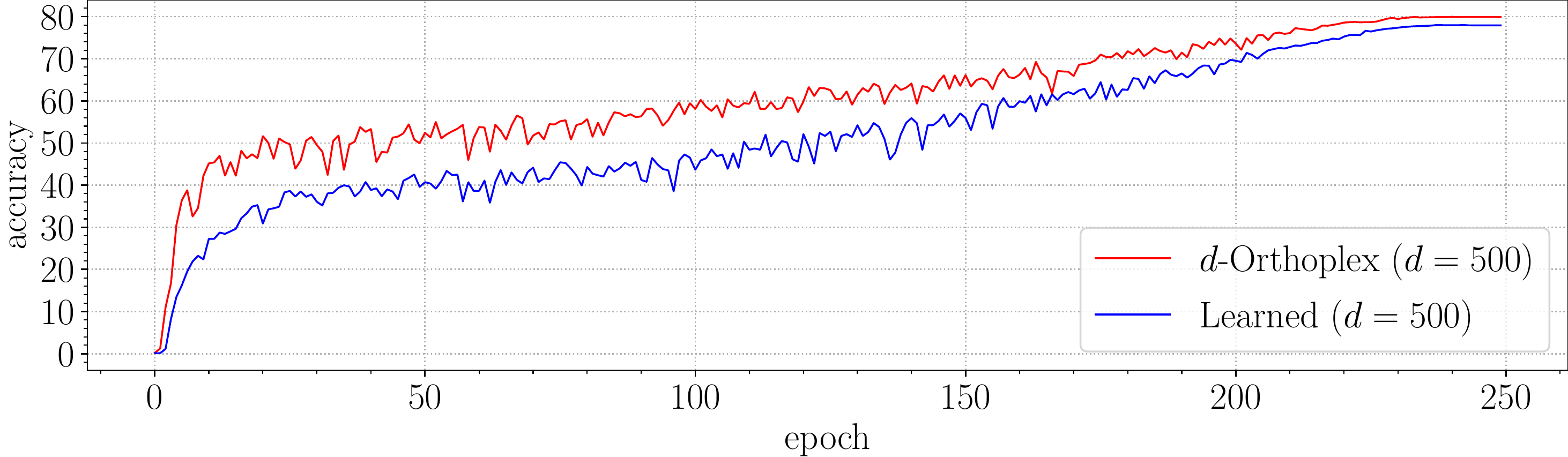}
    \vspace{-.1cm}
    }
    \subfigure[]{
        \hspace{-0.5cm}
        \includegraphics[width=0.49\textwidth]{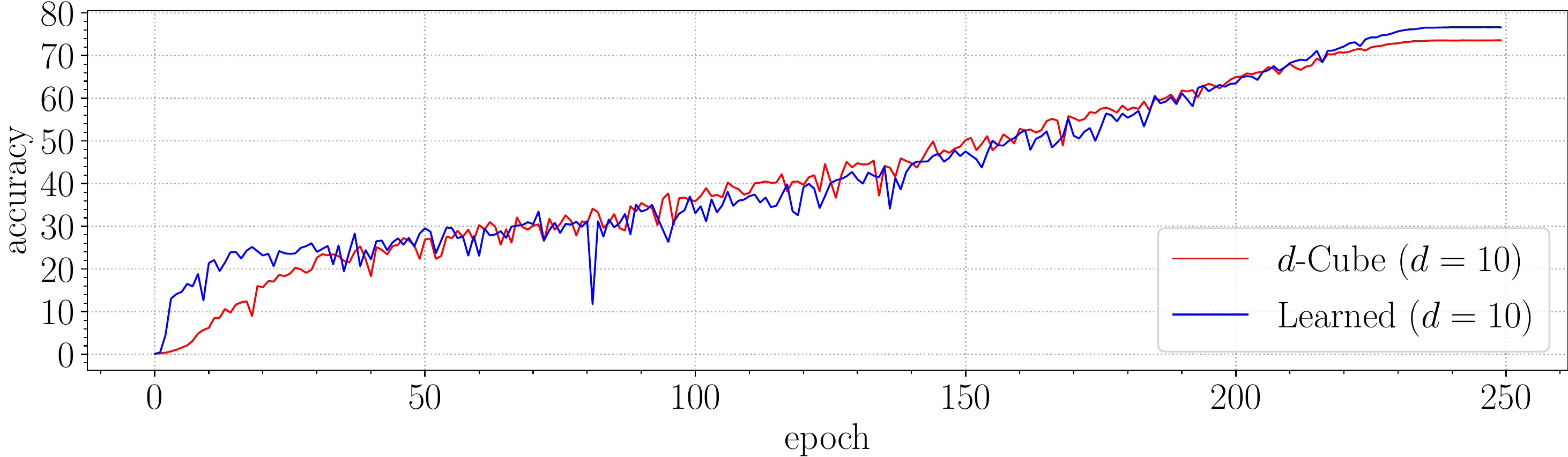}
    }
    \caption{Speed of convergence comparison (\textsc{Sgd+RandAug}). Test accuracy curves over the epochs on the ImageNet test set for the SeResNeXt50 architecture using the proposed fixed classifiers (red) and the standard trainable baselines (blue). \textit{(a):} The $d$-Simplex classifier, \textit{(b):} the $d$-Orthoplex classifier and \textit{(c):} the $d$-Cube classifer. The time to reach the same accuracy is shorter or equal for our method.}
    \label{fig:accuracy_seresnext}
\end{figure}

The time it takes to train a neural network model to address a classification problem is typically considered as the product of the training time per epoch and the number of epochs which need to be performed to reach the desired level of accuracy \cite{DBLP:conf/bigdataconf/JustusBBM18}.
Although in our case the training time per epoch is lower (the weights of the fixed classifier do not require back-propagation), it has a negligible effect due to the number of epochs required to reach a reasonable desired level of accuracy.
In Fig.~\ref{fig_learning_speed} and Fig.~\ref{fig:accuracy_seresnext} we report the classification accuracy over the epochs for 
the two different configurations of the training hyperparameters we evaluated.
\\
\indent
Specifically, Fig.~\ref{fig_learning_speed}(a)(\emph{top}) and Fig.~\ref{fig_learning_speed}(a)(\emph{bottom}) show the training error and the classification accuracy, respectively, over the epochs. The curves are obtained on the CIFAR100 dataset, using the VGG19 architecture and training is performed according to the Adam stochastic optimization.
Fig.~\ref{fig_learning_speed}(b) shows the accuracy curves of the proposed three fixed classifiers and the best performing learned classifier (i.e. $d=999$). The curves are obtained on the ImageNet dataset with the DenseNet169 architecture and learned according to the Adam optimizer. As can be noticed, $d$-Simplex and $d$-Orthoplex classifiers have lower or equal time to reach any desired level of accuracy than the learned and Hadamard fixed classifiers. The $d$-Cube classifier is the slowest and does not reach a comparable final performance. This is due to the different feature dimension ($d=10)$ and topology. However, when compared with a learned classifier with same feature dimension (as discussed in the next paragraph) the training time is similar.
\\
\indent
Fig.~\ref{fig:accuracy_seresnext}(a) and Fig.~\ref{fig:accuracy_seresnext}(b) show the time to reach accuracy using SGD+RandAug on SeResNeXt50 for the $d$-Simplex and $d$-Orthoplex (in red) fixed classifiers and the learned classifiers (in blue), respectively.
As evidenced in the figure, the learned classifiers require about 150 epochs to obtain the accuracy that the $d$-Simplex and $d$-Orthoplex achieve in 50 and 90 epochs, respectively. Although this gain reduces as  training progresses towards the end, our method achieves consistently better results.
Fig.~\ref{fig:accuracy_seresnext}(c)
shows that the $d$-Cube classifier requires similar training time with slightly lower final accuracy.
\\
\indent
The general behavior of the curves shown in  Fig.~\ref{fig_learning_speed} and Fig.~\ref{fig:accuracy_seresnext}
is consistent across combinations of datasets, architectures, classifiers and training strategies. The training time to reach the same accuracy is shorter or equal for our method and the time reduction follows the complexity of the embeddings defined by each regular polytope fixed classifier.

\indent
\\
\indent
Overall, we have demonstrated that Regular Polytope Networks provide a novel, effective and easy approach to fixed classifiers that achieves comparable state-of-the-art performance with the standard trainable classifiers. They provide faster speed of convergence and a significant reduction in model parameters. To facilitate replication of our experiments, the code will be made publicly available.

\section{Discussion: Potential and Challenges}
Our finding may have implications in those Deep Neural Network learning contexts in which a classifier must be robust against changes of the feature representation while learning. This is the case of incremental learning settings, especially when features are stored in memory banks while learning  \cite{DBLP:journals/corr/GravesWD14, DBLP:journals/corr/WestonCB14,graves2016hybrid}. Despite recent advances, methods inspired by memory-augmented deep neural networks are still limited when it comes to incremental learning. The method \cite{DBLP:conf/iclr/KaiserNRB17} simplifies the original fully differentiable end-to-end idea. Except for the nearest-neighbor query to the memory bank, their approach is fully differentiable, can be trained end-to-end and operates in a incremental manner (i.e. without the need of resetting during training). However, the features stored in the memory bank remain fixed (i.e. they are not undergoing learning) and only the memory bank is learned.
Our approach may have a promising potential for learning both the feature and the memory without considering their joint learning. The intuition is that every time the internal feature representation changes the memory bank must be relearned from scratch. Our method can mitigate the need of feature relearning by keeping the compatibility of features between learning steps thanks to their stationarity. Concurrent to this work, \cite{Shen_2020_CVPR} addresses a similar problem in terms of feature ``back-compatibility'' and exploits a pre-trained fixed classifier to avoid re-indexing a memory bank containing the gallery features of a retrieval system that has been updated. 
\\
\indent
This basic idea can be in principle applied to the many computer vision tasks that have benefited from memory based learning. Among them we mention \cite{PERNICI2020102983, Pernici_2018_CVPR, pernici2017unsupervised} for cumulative learning of face appearance models from video stream,  \cite{Beery_2020_CVPR, DBLP:conf/iccv/DengHSZXMRG19, DBLP:conf/iccv/ShvetsLB19, DBLP:conf/cvpr/WuF0HKG19, DBLP:conf/iccv/WuCWZ19} for object detection, \cite{DBLP:conf/iccv/OhLXK19} for video object segmentation and \cite{DBLP:conf/eccv/YangC18} for visual object tracking. The works \cite{Beery_2020_CVPR, DBLP:conf/iccv/DengHSZXMRG19, DBLP:conf/iccv/ShvetsLB19, DBLP:conf/cvpr/WuF0HKG19, DBLP:conf/iccv/WuCWZ19} accumulate context from pre-computed feature banks (with fixed pre-trained feature extractors i.e. not undergoing learning). The feature banks extend the time horizon of their network up to 60 second in \cite{DBLP:conf/cvpr/WuF0HKG19} or to one month in \cite{Beery_2020_CVPR} and achieve strong results on spatiotemporal localization. The works \cite{PERNICI2020102983, Pernici_2018_CVPR, pernici2017unsupervised} accumulate extracted face features in a memory bank to preserve all the past knowledge without forgetting and at the same time handle the non-stationarity of the data stream. 
At a high level, all these approaches can be framed as a non-parametric estimation method (like nearest neighbors) sitting on top of a high-powered parametric function (Faster R-CNN in the case of object detection \cite{Beery_2020_CVPR}, a face feature extractor in \cite{PERNICI2020102983} and \cite{Pernici_2018_CVPR}, a SiamFC feature extractor  \cite{DBLP:conf/eccv/BertinettoVHVT16} for object tracking in \cite{DBLP:conf/eccv/YangC18}). 
These methods use a fixed representation that is not incrementally learned as it would require re-encoding all the images in the memory bank. 
Avoiding re-encoding images can be advantageous in applications where images cannot be stored for privacy reasons (i.e. face recognition, applications in medical imaging, etc.).
Clearly, also Multi-Object Tracking \cite{CIAPARRONE202061,salvagnini2014information} can benefit from memory based learning.

\section{Conclusion}
We have shown that a special set of fixed classifiers based on regular polytopes generates stationary features by maximally exploiting the available representation space. The proposed method is simple to implement and theoretically correct.
Experimental results confirm both the theoretical analysis and the generalization capability of the approach across a range of datasets, baselines and architectures. Our RePoNet solution improves and generalizes the concept of a fixed classifier, recently proposed in \cite{hoffer2018fix}, to a larger class of fixed classifier models exploiting the inherent symmetry of regular polytopes in the feature space.

Our findings may have implications in all of those Deep Neural Network learning contexts in which a classifier must be robust against changes of the feature representation while learning  
as in incremental and continual learning settings.

\appendix
\subsection*{Computing the Angle Between Adjacent Classifier Weights}

The angle between a vertex and its adjacent vertices in a regular polytope can be computed following the same mathematical formulation used to compute its dihedral angle.
The dihedral angle of a regular $d$-Simplex is the acute angle formed by a pair of intersecting faces. In the case $d = 2$ the dihedral angle is the angle at the vertex of an equilateral triangle, while in the case $d = 3$ is the angle formed by the faces of the regular tetrahedron.

Because the dual polytope of a regular $d$-Simplex is also a regular $d$-Simplex, the angle $\theta$ between pairs of vertices can be expressed as \cite{parks2002elementary}:
\begin{equation}
\theta = \pi - \delta,  
\label{relationship_dihedral}
\end{equation}
where $\delta$ is the dihedral angle.
Since the dihedral angle of a regular $d$-Simplex is known to be \cite{parks2002elementary}\cite{coxeter1963regular}: 
\begin{equation}
\delta = \arccos{ \Big ( \frac{1}{d} \Big ) }
,
\label{dihedral_simplex}
\end{equation}
substituting Eq.~\ref{dihedral_simplex} into Eq.~\ref{relationship_dihedral} we obtain:
$$
\theta = \pi -  \arccos{ \Big ( \frac{1}{d} \Big )}
$$
which simplifies to: 
$$
\theta = \arccos{\Big (- \frac{1}{d} \Big)}.
$$   
The Eq. above provides the value between any pair of vectors in a $d$-Simplex.  

The calculation for the the $d$-Cube follows a similar argument.
The dihedral angle of a regular $d$-Orthoplex is known to be $\arccos{((2-d)/d)}$. Since the $d$-Cube is the dual of the $d$-Orthoplex the angle defined by a vertex of a $d$-Cube and its adjacent vertices is:
\begin{equation}
\theta = \arccos{ \Big ( \frac{d-2}{d} \Big ) }.
\end{equation}

\IEEEpeerreviewmaketitle

\section*{Acknowledgment}
This work was partially supported by the European Commission under European Horizon 2020 Programme, grant number 951911 - AI4Media.
This research was also partially supported by NVIDIA Corporation with the donation of Titan Xp GPUs, Leonardo Finmeccanica S.p.A., Italy.  We thank Francesco Calabrò (Leonardo), Giuseppe Fiameni (Nvidia) and Marco Rorro (CINECA) for their support.

\ifCLASSOPTIONcaptionsoff
  \newpage
\fi

\bibliographystyle{IEEEtran}
\bibliography{bib.bib}

\vfill\eject

\begin{IEEEbiography}
[{\includegraphics[width=1in,height=1.25in,clip,keepaspectratio]{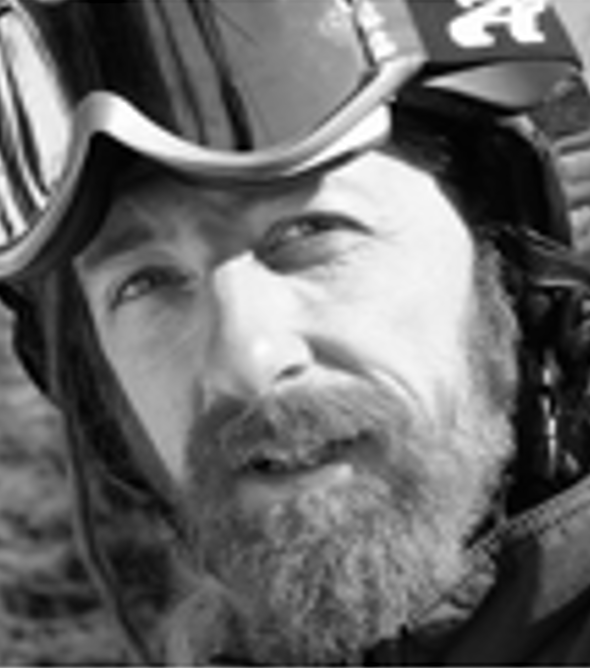}}]{Federico Pernici}
Federico Pernici received the laurea degree in Information Engineering in 2002, the post-laurea degree in Internet Engineering in 2003 and the Ph.D. in Information and Telecommunication Engineering in 2005 from the University of Firenze, Italy. Since 2002 he has been a research assistant at MICC Media Integration and Communication Center, assistant professor and adjunct professor at the University of Firenze. His scientific interests are computer vision and machine learning with a focus on different aspects of
visual tracking, incremental learning and representation learning. Presently, he is Associate Editor of Machine Vision and Applications journal. 
\end{IEEEbiography}
\vskip 0pt plus -1fil
\begin{IEEEbiography}
   [{\includegraphics[width=1in,height=1.25in,keepaspectratio]{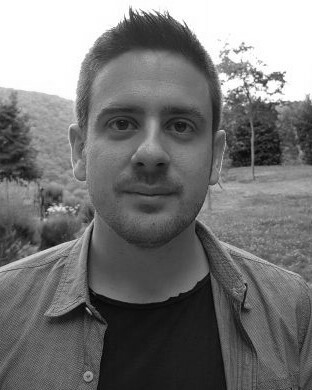}}]{Matteo Bruni}
Matteo Bruni received the M.S. degree (cum laude) in Computer Engineering 
from the University of Firenze, Italy, in 2016. Presently, he is a Ph.D. student at the University of Firenze at MICC, Media Integration and Communication Center, University of Firenze. His research interests include pattern recognition and computer vision with specific focus on feature embedding, face recognition and incremental learning.
\end{IEEEbiography}
\vskip 0pt plus -1fil
\begin{IEEEbiography}
[{\includegraphics[width=1in,height=1.25in,clip,keepaspectratio]{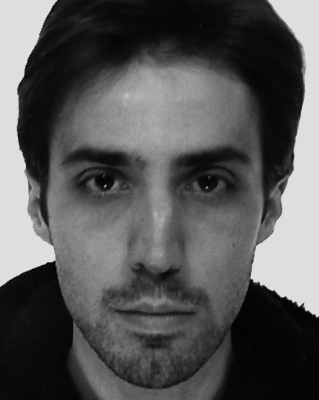}}]{Claudio Baecchi}
Claudio Baecchi received his Master's Degree in computer engineering from University of Florence in 2013 and the Ph.D in 2017. 
Currently he is working at the Visual Information and Media Lab at Media Integration and Communication Centre, University of Florence.
His current researches in the computer vision field cover sentiment and polarity classification in web videos, face and emotion recognition.
\end{IEEEbiography}
\vskip 0pt plus -1fil
\begin{IEEEbiography}
   [{\includegraphics[width=1in,height=1.25in,clip,keepaspectratio]{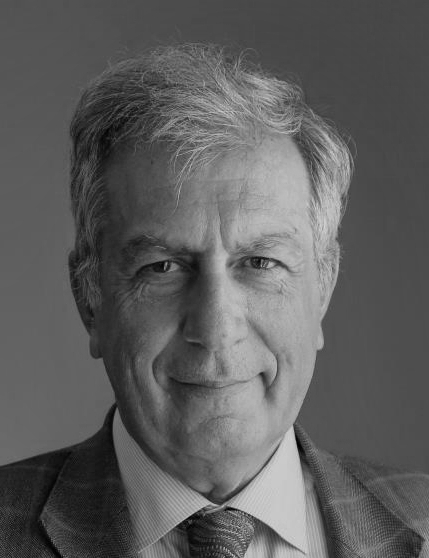}}]{Alberto Del Bimbo}
Prof. Del Bimbo is Full Professor at the University of Firenze, Italy and the Director of MICC Media Integration and Communication Center. He is the author of over 350 scientific publications in computer vision and multimedia and principal investigator of technology transfer projects with industry and governments. He was the Program Chair of ICPR 2012, ICPR 2016 and ACM Multimedia 2008, and the General Chair of IEEE ICMCS 1999, ACM Multimedia 2010, ICMR 2011 and ECCV 2012. He is the General Chair of the forthcoming ICPR 2020. He is the Editor in Chief of ACM TOMM Transactions on Multimedia Computing Communications and Applications and Associate Editor of Multimedia Tools and Applications and Pattern Analysis and Applications journals. He was Associate Editor of IEEE Transactions on Pattern Analysis and Machine Intelligence, IEEE Transactions on Multimedia and Pattern Recognition and also served as the Guest Editor of many Special Issues in highly ranked journals. Prof. Del Bimbo is IAPR Fellow and ACM Distinguished Scientist and is the recipient of the 2016 ACM SIGMM Award for \emph{Outstanding Technical Contributions to Multimedia Computing Communications and Applications}.
\end{IEEEbiography}

\end{document}